\documentclass{article}

\usepackage[preprint]{neurips_2026}

\usepackage[utf8]{inputenc} 
\usepackage[T1]{fontenc}    
\usepackage{hyperref}       
\usepackage{url}            
\usepackage{booktabs}       
\usepackage{amsfonts}       
\usepackage{nicefrac}       
\usepackage{microtype}      
\usepackage{xcolor}         

\usepackage{graphicx}
\usepackage{subcaption}
\usepackage{amsmath}
\usepackage{amssymb}
\usepackage{mathtools}
\usepackage{amsthm}

\usepackage{bbm}          
\usepackage{enumitem}     
\usepackage[table]{xcolor}

\usepackage[capitalize]{cleveref}
\crefname{equation}{Eq.}{Eqs.}
\Crefname{equation}{Eq.}{Eqs.}
\crefname{section}{Sec.}{Secs.}
\Crefname{section}{Sec.}{Secs.}
\crefname{subsection}{Sec.}{Secs.}
\Crefname{subsection}{Sec.}{Secs.}
\crefname{subsubsection}{Sec.}{Secs.}
\Crefname{subsubsection}{Sec.}{Secs.}
\crefname{appendix}{\S}{\S}
\Crefname{appendix}{\S}{\S}
\crefname{table}{Tab.}{Tabs.}
\Crefname{table}{Tab.}{Tabs.}
\crefname{figure}{Fig.}{Figs.}
\Crefname{figure}{Fig.}{Figs.}
\usepackage{wrapfig}
\usepackage{caption}
\usepackage{algorithm}
\usepackage{algorithmic}

\theoremstyle{plain}
\newtheorem{theorem}{Theorem}[section]
\newtheorem{proposition}[theorem]{Proposition}

\theoremstyle{definition}

\theoremstyle{remark}

\usepackage[textsize=tiny]{todonotes}

\DeclareMathOperator*{\argmin}{arg\,min}
\newcommand{\Equality}{E_{\text{quality}}}
\newcommand{\Econstraint}{E_{\text{constraint}}}
\newcommand{\muq}{\mu_{\text{quality}}}
\newcommand{\sigmaq}{\sigma_{\text{quality}}}

\title{Distributional Energy-Based Models for \\
  Uncertainty-Aware Structured LLM Reasoning}

\author{                                            \textbf{Shireen Kudukkil Manchingal}\thanks{Corresponding author: \texttt{shireen.manchingal@oxdynamics.com}}\hspace{2.7em}%
\textbf{Abhey Kalia}\hspace{2.7em}%
\textbf{Fernanda Gonçalves}\\[5pt]                  
\textbf{Shebin Rawther}\\[5pt]
Oxford Dynamics\\[5pt]                              Harwell Science and Innovation Campus, United Kingdom
  }

\begin{document}

\maketitle

\begin{abstract}
\vspace{-5pt}
When Large Language Models produce structured outputs such as travel plans, code solutions, or multi-step proofs, individual reasoning steps may appear correct while the output as a whole violates budgets, fails test cases, or contradicts earlier deductions.
We propose a decomposed energy function that combines a learned quality scorer with deterministic analytical constraint penalties for verifying structured LLM outputs.
The quality scorer is a heterogeneous ensemble of low-rank adapters on a single frozen encoder (3\% trainable parameters); the ensemble mean ranks candidates while the standard deviation quantifies epistemic uncertainty, driving a two-pass inference loop that triggers targeted regeneration or abstention.
Across five benchmarks (GSM8K, MuSR, TravelPlanner, TACO, Knights \& Knaves), our 149M-parameter verifier orchestrating a pool of 7--26B open generators outperforms single-shot Qwen-72B on every benchmark, matches Claude Sonnet~4.6 on MuSR (67.7\% vs.\ 68.0\%), and reduces constraint violations by 53\% relative to Opus~4.6 on TravelPlanner (oracle 0.028, random 0.231).
The two routes are complementary: structural verification wins when constraints are checkable (the verifier captures signal frontier models cannot self-detect), while pretraining-scale priors win where they are not (narrative inference, code semantics).
A cross-dataset confounding analysis confirms genuine quality discrimination on four reasoning tasks and identifies a model-identity shortcut on code, mitigated via last-layer retraining.
Scorers trained on difficult data transfer zero-shot: a MuSR-trained scorer achieves 93.9\% on GSM8K without seeing a math problem.
\end{abstract}

\vspace{-10pt}
\section{Introduction}
\label{sec:intro}
\vspace{-3pt}

Large Language Models have advanced rapidly, achieving strong performance across translation, summarisation, code generation, and open-ended dialogue~\citep{brown_2020_gpt3, openai_2023_gpt4}.
A central driver of this progress is reasoning: the ability to decompose a problem into intermediate steps and arrive at a justified conclusion.
Chain-of-thought (CoT) prompting~\citep{wei_2022_cot, kojima_2022_zero_shot_cot} unlocked this capability by eliciting step-by-step reasoning traces, and subsequent work has extended it through self-consistency~\citep{wang_2023_selfconsistency}, least-to-most decomposition~\citep{zhou_2023_leasttomost}, and tree-structured search~\citep{yao_2023_tot}.

\begin{wrapfigure}{r}{0.45\textwidth}
    \vspace{-15pt}
    \hspace{-0.3cm}
\includegraphics[width=0.5\textwidth]{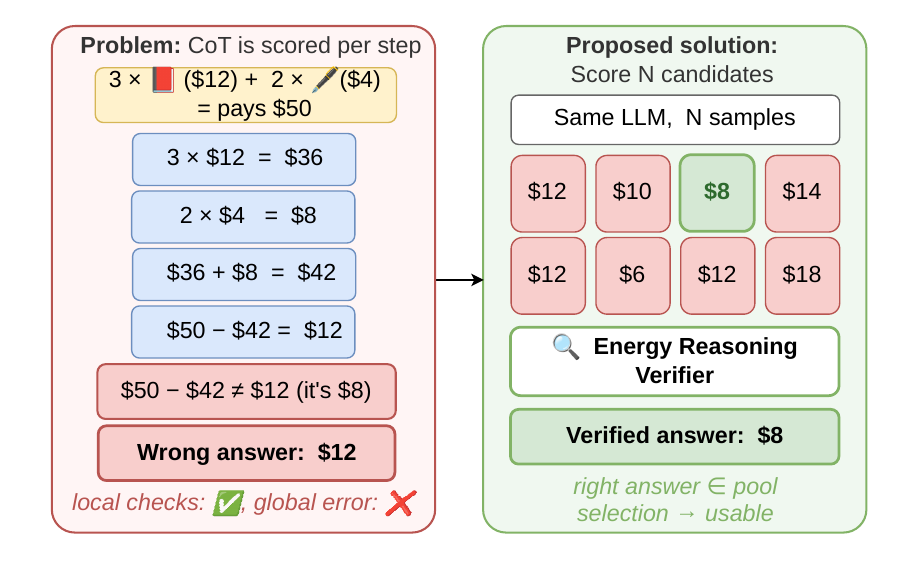}
    \vspace{-15pt}
    \caption{\textbf{Left:} Chain-of-thought scores each step locally; cross-step violations are invisible. \textbf{Right:} Distributional energy scores $N$ candidates globally, verifies constraints analytically, and quantifies confidence via ensemble agreement.}
    \label{fig:motivation}
    \vspace{-12pt}
\end{wrapfigure}

Yet reasoning in current models remains fundamentally limited by the autoregressive generation mechanism: each token is committed in sequence, with no global scoring function to evaluate whether the output as a whole is consistent. Local correctness at each step does not compose into global correctness for structured outputs: cross-step violations are aggregation errors that exist only at the whole-output level (a budget exceeded only when all days are summed, a return-type mismatch propagating between functions, a single arithmetic error invalidating the rest of a derivation), and per-step scoring cannot detect them by construction, no matter how accurate.
Each step receives a local correctness signal yet cross-step violations remain invisible (\cref{fig:motivation}, left), so the wrong plan is delivered with no warning.
GPT-4-Turbo satisfies all constraints on TravelPlanner only 4.4\% of the time~\citep{xie_2024_travelplanner}, and even frontier models frequently produce code that fails on edge cases in competitive programming benchmarks~\citep{chen_2021_codex, li_2023_taco}.

The bottleneck, however, is not generation but \emph{selection}: pass@$N$ substantially exceeds pass@1, the correct candidate is usually somewhere in the pool of $N$ samples from the same LLM, yet existing scorers cannot reliably identify it~\citep{brown_2024_monkeys, stiennon_2020_summarize}.
Process reward models~\citep{lightman_2023_lets_verify, khalifa_2025_thinkprm} score individual reasoning steps rather than complete outputs.
Energy-based models (EBMs) offer a more natural framework for whole-output scoring, assigning low energy to good candidates and high energy to bad ones~\citep{lecun_2006_ebm_tutorial}, and recent work applies this principle to rerank reasoning traces~\citep{jiang_2025_eorm, chen_2025_ebmcot, tan_2025_rl_ebm}.
Yet existing methods share two limitations: (i)~they produce point-estimate scores with no indication of reliability~\citep{gleave_2022_uncertainty_rm}, offering no signal for when to trust, regenerate, or abstain; and (ii)~they operate on flat reasoning chains with no mechanism for verifying domain-specific constraints exactly.

In this paper, we show that this selection problem can be addressed with a small specialised verifier rather than by scaling the generator. We propose a \emph{decomposed, distributional} energy function $E = \muq + \lambda \cdot \Econstraint$ (\cref{fig:motivation}, right) combining a learned quality term $\muq$ with deterministic analytical constraint penalties $\Econstraint$. The same compact architecture works across five fundamentally different reasoning tasks (math, narrative inference, multi-constraint planning, executable code, formal logic).
The quality term $\muq$ is realised as a heterogeneous ensemble of $K$ low-rank adapters (LoRA;~\citealp{hu_2022_lora}) on a single frozen encoder, structurally diverse so that ensemble disagreement reflects genuine epistemic uncertainty rather than superficial variation from random seeds~\citep{fort_2019_deep_ensembles, lakshminarayanan_2017_simple_ensembles, gal_2016_dropout}.
The ensemble mean ranks candidates; the standard deviation $\sigmaq$ drives a two-pass inference loop that triggers regeneration with structured feedback under uncertainty or constraint violations, and abstains in the extreme case.
Crucially, the verifier is architecturally independent of the candidate generators, so its judgement does not collapse when the generators' self-reported reasoning is wrong.

The following are the main \textbf{contributions} of this paper:
\begin{itemize}[noitemsep, leftmargin=*, topsep=0pt, partopsep=0pt]
    \item \textbf{A novel modular energy decomposition with zero-shot transfer.}
    A transferable learned quality scorer $\Equality$ and swappable deterministic penalties $\Econstraint$: $\Equality$ wins beyond pretraining-scale priors, $\Econstraint$ wins on checkable constraints. The scorer transfers zero-shot (MuSR-trained reaches 93.9\% on GSM8K without seeing math); new tasks need only new constraint functions.
    \item \textbf{A novel distributional energy via heterogeneous LoRA ensemble.}
    A parameter-efficient ensemble of structurally diverse low-rank adapters produces an energy \emph{distribution} rather than a point estimate, enabling confounding diagnostics and $\sigma$-gated abstention.
    \item \textbf{Honest confounding analysis with causal verification.}
    On 4/5 tasks the scorer learns content-based quality with no model-identity shortcut (cross-model energy spreads $<0.25$, picks near-uniform across generators); on TACO, a causal style-swap test exposes a formatting shortcut, fixed by last-layer retraining~\citep{kirichenko_2023_dfr} at no accuracy cost. Multi-model candidate diversity substantially improves selection.
\end{itemize}
\textbf{Paper outline.} \cref{sec:related} reviews chain-of-thought reasoning, energy-based models, reward modelling, and test-time compute. \cref{sec:method} presents the energy decomposition (\ref{sec:method:formulation}), the distributional quality scorer with formal propositions (\ref{sec:method:quality}), training (\ref{sec:method:training}), analytical constraint penalties (\ref{sec:method:constraints}), and two-pass $\sigma$-triage (\ref{sec:method:twopass}). \cref{sec:experiments} reports the experimental setup (\ref{sec:exp:setup}) and results across all five benchmarks (\ref{sec:exp:results}). \cref{sec:conclusion} distils lessons learned and outlines future directions. The appendix provides proofs (\S\ref{app:proofs}), model and implementation details (\S\ref{app:models},~\S\ref{app:implementation}), dataset details and per-model candidate quality (\S\ref{app:training_data},~\S\ref{app:datasets}), and extended results: ablations (\S\ref{app:ablations}), energy diagnostics (\S\ref{app:energy}), per-difficulty breakdowns (\S\ref{app:difficulty}), selective prediction (\S\ref{app:selective}), two-pass analysis (\S\ref{app:twopass}), per-constraint frontier comparison (\S\ref{app:travelplanner_frontier}), the full transfer matrix (\S\ref{app:transfer}), confounding analysis (\S\ref{app:confounding}), and broader impact (\S\ref{app:impact}).

\vspace{-5pt}
\section{Related Work}
\label{sec:related}
\vspace{-5pt}

Chain-of-thought prompting~\citep{wei_2022_cot, kojima_2022_zero_shot_cot} elicits step-by-step reasoning; self-consistency~\citep{wang_2023_selfconsistency}, least-to-most prompting~\citep{zhou_2023_leasttomost}, and Tree of Thoughts~\citep{yao_2023_tot} extend it via marginalisation, decomposition, and branching search. Process reward models give finer-grained supervision: \citet{uesato_2022_prm} and \citet{lightman_2023_lets_verify} show step-level outperforms outcome-level verification, and \citet{khalifa_2025_thinkprm} train models that generate reasoning traces before scoring, with a fraction of process labels. All produce point-estimate scores on flat reasoning chains, with no mechanism for verifying that structured multi-section outputs satisfy global constraints jointly.

Energy-based models assign scalar energies to input-output pairs~\citep{lecun_2006_ebm_tutorial} and have been applied to text generation~\citep{deng_2020_residual_ebm}; \citet{blondel_2025_arms_ebms} prove autoregressive language models are themselves EBMs, and \citet{carbone_2024_ebm_hitchhiker} survey the broader landscape. For reasoning, \citet{jiang_2025_eorm} rerank CoT on GSM8K/MATH (90.7\%, 55M params); \citet{chen_2025_ebmcot} apply energy-based calibration to implicit CoT; \citet{tan_2025_rl_ebm} connect RL-tuned language models to EBMs theoretically. Across this work, energy is a deterministic scalar with no reliability signal; \citet{gleave_2022_uncertainty_rm} highlight this gap for reward models but use standard ensemble variance rather than structurally heterogeneous adapters. Our pairwise-preference learning connects to RLHF~\citep{christiano_2017_rlhf, ouyang_2022_instructgpt} and DPO~\citep{rafailov_2023_dpo} but targets post-hoc verification of complete outputs; we further produce a \emph{distributional} energy via a heterogeneous ensemble, decompose energy into learned and analytical components, and target structured multi-section outputs.
A complementary line studies test-time compute allocation. \citet{snell_2024_scaling_testtime} show verifier-guided search can outperform scaling model parameters, \citet{deepseek_2025_r1} show reinforcement learning alone can incentivise advanced reasoning, and \citet{brown_2024_monkeys} identify candidate \emph{selection} as the key bottleneck: pass@$N$ exceeds pass@1 but existing methods close the gap only partially. For structured planning, \citet{lee_2025_mind_evolution} reach 95.6\% on TravelPlanner via evolutionary search with Gemini 1.5 Flash, \citet{xia_2025_agentrm} use reward modelling for agent generalisation, and \citet{wang_2024_moa} combine multiple LLMs via mixture-of-agents. We address the selection bottleneck via energy-based reranking and the verification gap via analytical constraint penalties, using ensemble uncertainty to decide when selection is reliable.

\vspace{-5pt}
\section{Methodology}
\label{sec:method}
\vspace{-5pt}

\begin{figure}[!ht]
    \centering
    \includegraphics[width=\textwidth]{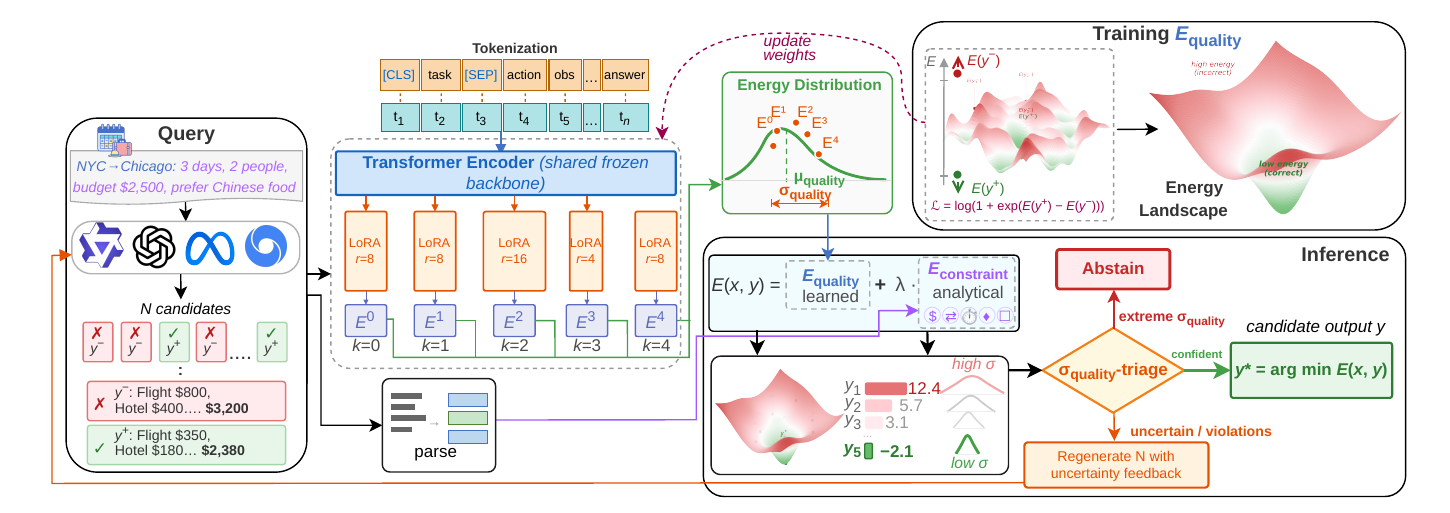}
    \vspace{-15pt}
    \caption{Method overview. \textbf{Top:} Training. Correct ($y^+$) / incorrect ($y^-$) candidate pairs pass through the frozen ModernBERT backbone with one of $K{=}5$ heterogeneous LoRA adapters; each adapter's MLP head maps \texttt{[CLS]} to a scalar $E^k$ trained via Bradley-Terry. The $K$ energies form a distribution: $\muq$ ranks, $\sigmaq$ quantifies confidence. \textbf{Bottom:} Inference. Four LLMs generate $N{=}32$ candidates, scored in parallel by the ensemble and by analytical constraints ($\Econstraint$), ranked by $E = \muq + \lambda \cdot \Econstraint$, then $\sigmaq$-triaged.}
    \label{fig:method}
    \vspace{-10pt}
\end{figure}

We propose a Distributional EBM (\cref{fig:method}) with two stages: a quality scorer trained on contrastive candidate pairs (top), and inference-time candidate ranking with $\sigma$-triaged regeneration (bottom).

\vspace{-6pt}
\subsection{Problem Formulation}
\label{sec:method:formulation}
\vspace{-3pt}

Let $x$ denote an input problem and let $\{y_1, \ldots, y_N\}$ denote $N$ candidate structured outputs generated by frozen base LLMs via temperature sampling.
We propose the decomposed energy function:
\begin{equation}
    E(x, y) = \muq(x, y) + \lambda \cdot \Econstraint(x, y),
    \label{eq:energy}
\end{equation}
and the selected candidate is the energy minimiser over the candidate pool $\mathcal{Y} = \{y_1, \ldots, y_N\}$, $y^{*} = \argmin_{y \in \mathcal{Y}} E(x, y)$, where $\muq(x, y)$ is the ensemble mean energy from the learned quality scorer, $\Econstraint(x, y)$ is a sum of analytical penalty terms, and $\lambda > 0$ controls their relative weight.
The two components serve complementary roles: $\Equality$ captures general reasoning quality learned from training data, while $\Econstraint$ encodes task-specific rules that can be verified exactly without any learning.

\vspace{-6pt}
\subsection{Distributional Quality Scorer}
\label{sec:method:quality}
\vspace{-3pt}

We instantiate the quality scorer as an ensemble of $K$ EBMs sharing a frozen pretrained bidirectional Transformer encoder~\citep{devlin_2019_bert, warner_2024_modernbert}. We choose an encoder over the now-standard causal-LM verifier so that a single \texttt{[CLS]} forward pass scores the whole \texttt{[problem]||[candidate]} pair jointly with bidirectional attention, at one to three orders of magnitude fewer parameters than CLM-based reward models or PRMs.
To each member $k$ we attach a lightweight low-rank adapter~\citep{houlsby_2019_adapters, hu_2022_lora} that modifies a subset of the encoder's attention matrices, together with an independent energy head.
Given a problem $x$ and candidate $y$, the input is tokenised as $\text{input}_k = \texttt{[CLS]}\ x\ \texttt{[SEP]}\ y\ \texttt{[SEP]}$.
The frozen encoder, augmented by adapter $k$, produces contextualised representations.
The \texttt{[CLS]} representation $\mathbf{h}^k_{\text{CLS}}$ is passed through an MLP energy head producing a scalar $E^k_{\text{quality}}(x, y)$ (architecture in \S\ref{app:implementation}).
Each adapter and energy head is trained independently; the frozen backbone is shared.
The distributional output for candidate $y$ is the ensemble mean $\muq(x, y) = \tfrac{1}{K} \sum_{k=1}^{K} E^k_{\text{quality}}(x, y)$ and standard deviation $\sigmaq(x, y) = \text{std}(\{E^1_{\text{quality}}, \ldots, E^K_{\text{quality}}\})$.
The mean $\muq$ is used for ranking (\cref{eq:energy}); the standard deviation $\sigmaq$ drives the two-pass inference loop (\cref{sec:method:twopass}) and provides a calibration signal for post-hoc analysis.

Diversity comes from three sources: \emph{structural heterogeneity} (the $K$ adapters differ in rank, scaling factor, and target attention modules; \cref{tab:ensemble_arch} in \S\ref{app:implementation}), so disagreement reflects genuine epistemic uncertainty rather than seed variation~\citep{fort_2019_deep_ensembles, lakshminarayanan_2017_simple_ensembles}; independent random initialisation per adapter; and 80\% data bagging~\citep{breiman_1996_bagging}. Trainable parameters per member ($\sim$3\% of backbone) consist only of the LoRA weights and the energy head.

\textbf{Theoretical justification.}
A natural question is what the distributional energy ($\muq$, $\sigmaq$) provides beyond a point-estimate scorer.
We present two results that together justify the combination of decomposition and ensemble.

\begin{proposition}[Ensemble ranking under correlated adapters]
\label{prop:condorcet}
Let each adapter $k$ correctly rank a random (correct, incorrect) pair with probability $q > \tfrac{1}{2}$, and let $\rho \in [0, 1]$ denote the pairwise rank correlation between adapters.
Under a Gaussian approximation for $K \geq 5$, the probability that the ensemble mean $\muq$ correctly ranks the pair, and its $K\to\infty$ limit (for any $\rho > 0$), are:
\vspace{-3pt}
\begin{equation}
    P_{\emph{ensemble}}(K, q, \rho) = \Phi\!\left(\tfrac{\sqrt{K}\,(q - \tfrac{1}{2})}{\sqrt{q(1{-}q)(1 + (K{-}1)\rho)}}\right), \quad
    P^\infty(q, \rho) = \Phi\!\left(\tfrac{q - \tfrac{1}{2}}{\sqrt{q(1{-}q)\,\rho}}\right) < 1,
    \label{eq:condorcet}
\end{equation}
where $\Phi$ is the standard normal CDF and $P^\infty < 1$ is a finite ceiling that no ensemble size can surpass unless $\rho = 0$.
\end{proposition}
\vspace{-3pt}

Three consequences follow (proof and detailed analysis in \S\ref{app:proofs}). $\partial P / \partial \rho < 0$, so structurally heterogeneous adapters (lower $\rho$) strictly outperform same-architecture ensembles; when $\rho = 1$, $P_{\text{ensemble}} \to q$ regardless of $K$; and for any $\rho > 0$ the ceiling in \cref{eq:condorcet} is binding for moderate $q$ (e.g., narrative reasoning) and slack only when $q$ is near unity (e.g., K\&K).
\vspace{-3pt}

\Cref{prop:condorcet} justifies the ensemble for ranking.
The following result justifies the \emph{decomposition} for uncertainty: because $\Econstraint$ is deterministic, $\sigmaq$ reflects purely the epistemic uncertainty of the learned quality component, uncontaminated by constraint uncertainty.

\begin{proposition}[Partitioned variance under energy decomposition]
\label{prop:variance}
Let $E(x, y) = \muq(x, y) + \lambda \cdot \Econstraint(x, y)$ where $\Econstraint$ is a deterministic analytical function.
The predictive variance of the total energy decomposes as:
\begin{equation}
    \mathrm{Var}(E) = \mathrm{Var}(\Equality) + \underbrace{\mathrm{Var}(\Econstraint)}_{=\,0} + \underbrace{2\,\mathrm{Cov}(\Equality, \Econstraint)}_{=\,0} = \sigmaq^2.
    \label{eq:variance_decomp}
\end{equation}
For a monolithic scorer $E_{\text{mono}}$ that jointly learns both components, the predictive variance is $\mathrm{Var}(\Equality^{\text{mono}}) + \mathrm{Var}(\Econstraint^{\text{mono}}) + 2\,\mathrm{Cov}(\Equality^{\text{mono}}, \Econstraint^{\text{mono}})$, which equals $\sigmaq^2$ only in the degenerate case that $\Econstraint^{\text{mono}}$ is learned exactly; otherwise the variance mixes constraint-prediction error into the uncertainty signal and is no longer interpretable as quality uncertainty.
\end{proposition}
\vspace{-3pt}

This makes $\sigmaq$ an \emph{interpretable} uncertainty signal (quality only, with zero contribution from constraint verification), and motivates the $\sigma$-triage decision rule (\cref{sec:method:twopass}): accept when $\sigmaq$ and $\Econstraint$ are both low, regenerate when either is high, abstain when $\sigmaq$ is extreme. Proof and full implications are in \S\ref{app:proofs}.

\vspace{-5pt}
\subsection{Training}
\label{sec:method:training}
\vspace{-5pt}

We train each member independently with the Bradley--Terry pairwise loss~\citep{bradley_1952_paired} on bagged 80\% subsets $\mathcal{D}^k$ of training problems, where $Y^+_x$ and $Y^-_x$ are the correct/incorrect outputs for $x$.
The objective for member $k$ is:
\vspace{-3pt}
\begin{equation}
    \mathcal{L}^{k} = \frac{1}{|\mathcal{D}^k|}\sum_{x \in \mathcal{D}^k}\frac{1}{|P_x|}\sum_{(y^+, y^-) \in P_x} \log\!\Big(1 + \exp\!\big(E^k_{\text{quality}}(x, y^+) - E^k_{\text{quality}}(x, y^-)\big)\Big),
    \label{eq:bt_loss}
\end{equation}
where $P_x \subset Y^+_x \times Y^-_x$ is a set of contrastive pairs sampled per problem, capped at a maximum count to prevent easy problems from dominating the gradient.
Members are optimised separately ($\mathcal{L}^1, \ldots, \mathcal{L}^K$); no parameters are shared across members beyond the frozen backbone.
The loss pushes the energy of correct outputs below incorrect ones; the logistic form saturates naturally for well-separated pairs, avoiding the need for explicit energy regularisation.

Contrastive pairs are split at the \emph{problem level} to prevent data leakage; labelling and hyperparameters are in \cref{sec:exp:setup}.
The trained ensemble is applied \textbf{zero-shot} to out-of-domain tasks; task-specific verification is handled entirely by $\Econstraint$.

\vspace{-5pt}
\subsection{Analytical Constraint Penalties}
\label{sec:method:constraints}
\vspace{-5pt}

We define $\Econstraint$ as a sum of analytical penalty terms, each encoding a specific task constraint: $\Econstraint(x, y) = \sum_{j} w_j \cdot C_j(x, y)$, where $C_j$ is the penalty for constraint $j$ and $w_j$ is its weight.
Each $C_j$ returns 0 when the constraint is satisfied and a positive value proportional to the degree of violation.
The constraint terms are deterministic: each is a direct computation over the parsed structured output with no learned parameters.
They are also modular: adding or removing a constraint changes $\Econstraint$ without affecting $\Equality$ or requiring retraining.
The specific constraint terms for each evaluation domain are defined in \S\ref{app:datasets}.

\vspace{-5pt}
\subsection{Two-Pass Inference with $\sigma$-Triage}
\label{sec:method:twopass}
\vspace{-5pt}

We extend single-pass reranking with a two-pass inference loop driven by ensemble uncertainty and constraint satisfaction, recovering when no good candidate exists or the scorer is uncertain (\cref{alg:inference} in \S\ref{app:twopass}).
Multiple frozen LLMs each generate $N_m$ candidates via temperature sampling, which are pooled into the first-pass set $\mathcal{Y}_1$ and scored.
Each candidate receives $\muq$ and $\sigmaq$ from the $K{=}5$ ensemble and $\Econstraint$ from the analytical constraint checker.
Let $y^{*}_1 = \argmin_{y \in \mathcal{Y}_1} E(x, y)$ denote the first-pass selection.
The system then takes one of three actions, governed by thresholds $\theta_\sigma < \theta_{\text{abstain}}$:
\begin{equation}
    \text{action}(y^{*}_1) =
    \begin{cases}
        \mathrm{accept}    & \text{if } \sigma \le \theta_\sigma \;\land\; C = 0, \\
        \mathrm{regenerate} & \text{if } \sigma \le \theta_a \;\lor\; C > 0, \\
        \mathrm{abstain}   & \text{if } \sigma > \theta_a,
    \end{cases}
    \label{eq:triage}
\end{equation}
where $\sigma = \sigmaq(y^*_1)$, $C = \Econstraint(y^*_1)$, and $\theta_a = \theta_{\text{abstain}}$.
On regenerate, constraint violations are formatted as natural-language feedback and used to draw $N_2$ additional candidates $\mathcal{Y}_2$.
The pooled set $\mathcal{Y} = \mathcal{Y}_1 \cup \mathcal{Y}_2$ is rescored, and the final selection is $y^{*} = \argmin_{y \in \mathcal{Y}} E(x, y)$, subject to a final abstention check on $\sigmaq(y^{*})$.

\vspace{-5pt}
\section{Experiments}
\label{sec:experiments}
\vspace{-5pt}

Our experiments (\Cref{sec:exp:results}) address five questions:
\textbf{(i)}~Is the bottleneck in structured LLM reasoning selection rather than generation?
\textbf{(ii)}~Does separating learned quality from analytical constraints improve correctness beyond either signal alone?
\textbf{(iii)}~Can ensemble disagreement reliably indicate when to trust, regenerate, or reject an output?
\textbf{(iv)}~Does the scorer learn genuine reasoning quality or exploit model-identity shortcuts?
\textbf{(v)}~Does the quality scorer transfer across reasoning domains without retraining?
Extended results are in the Appendix (\S\ref{app:extended}).

\vspace{-10pt}
\subsection{Experimental Setup}
\label{sec:exp:setup}
\vspace{-5pt}

\paragraph{Datasets.}
We evaluate across five benchmarks chosen for fundamentally different verification signals: math~(\textbf{GSM8K};~\citealp{cobbe_2021_gsm8k}), narrative inference~(\textbf{MuSR};~\citealp{sprague_2024_musr}), multi-constraint planning~(\textbf{TravelPlanner};~\citealp{xie_2024_travelplanner}), executable code~(\textbf{TACO};~\citealp{li_2023_taco}), and formal logic~(\textbf{K\&K};~\citealp{xie_2024_kk}). We evaluate across all five domains together, where prior reranking work~\citep{jiang_2025_eorm, lee_2025_mind_evolution} has typically focused on math or planning alone. GSM8K is partially contaminated~\citep{yang_2023_rethink_benchmark, oren_2024_contamination}, so we rely on MuSR (released 2024, post-dating all four generators' pretraining) as the uncontaminated headline; MuSR results are reported on the full test split (151 problems across all three subsets), with per-subset breakdowns in \S\ref{app:difficulty}. TravelPlanner contributes five analytical constraints (budget, connectivity, completeness, preferences, diversity) defining $\Econstraint$, verified against its sandbox database; K\&K's statement-consistency checker is a perfect verifier; TACO uses execution-based verification (we filter \texttt{VERY\_HARD} with 0\% solve rate). Primary metrics: pass@1 (single-sample correctness, the deployment-relevant figure) for GSM8K, MuSR, TACO, and K\&K; mean violation score for TravelPlanner, since the strict zero-violation pass@1 compresses methods into a 0--30\% band and hides the order-of-magnitude differences captured by the continuous score. Splits, EDA, per-model candidate quality, and examples are in \S\ref{app:datasets} (\cref{tab:dataset_overview,tab:examples}); training data statistics in \S\ref{app:training_data}.

\vspace{-2pt}
\textbf{Candidate generators.}
We use four frozen generators chosen for open-weight access via OpenRouter and parameter-count diversity (7--26B): Qwen-2.5-7B-Instruct~\citep{qwen_2024_qwen25}, LLaMA-3.1-8B-Instruct~\citep{grattafiori_2024_llama3}, Gemma-4-26B-IT~\citep{google_2026_gemma}, and GPT-OSS-20B~\citep{openai_2025_gptoss}. Each produces 8 candidates per problem via temperature sampling ($T{=}0.7$, top-$p{=}0.95$~\citep{holtzman_2020_nucleus}), for $N{=}32$ candidates per problem; all generation uses the OpenRouter API. Pass~2 adds $N_2{=}8$ candidates from Qwen-2.5-7B conditioned on constraint feedback.

\vspace{-2pt}
\textbf{Baselines.}
We compare against three groups: (1) heuristic selection from the candidate pool (Greedy at $T{=}0$, random selection, Self-consistency majority vote, and an Oracle upper bound); (2) state-of-the-art frontier models (Claude Opus~4.6 and Sonnet~4.6~\citep{anthropic_2026_claude}, Qwen-72B-Instruct~\citep{qwen_2024_qwen25}) which generate one deterministic answer at $T{=}0$ rather than selecting from a pool; and (3) external learned selection methods that score the same candidate pool as our scorer: EORM~\citep{jiang_2025_eorm} (the closest prior energy-based method), Skywork-Reward-LLaMA-3.1-8B-v0.2~\citep{liu_2024_skywork} (an 8B reward model), Math-Shepherd PRM~\citep{wang_2024_mathshepherd} (a 7B step-level Process Reward Model), and LLM-as-Judge~\citep{zheng_2023_llmjudge} using Claude Sonnet~4.6 (frontier model scoring each candidate 1-10). Every learned baseline and frontier model in this comparison is substantially larger than our 149M-parameter scorer: Skywork at 8B ($54\times$), Math-Shepherd PRM at 7B ($47\times$), Qwen-72B ($\sim$483$\times$), and the frontier Claude models are several hundred billion parameters, one to three orders of magnitude larger than ours.

\vspace{-2pt}
\textbf{Scorer backbone.}
The primary backbone is ModernBERT-base~\citep{warner_2024_modernbert} (149M parameters, 8{,}192 token context); the long context accommodates MuSR narratives (3--6K characters) and full TravelPlanner itineraries without truncation.
DeBERTa-v3-base~\citep{he_2023_debertav3} (86M parameters, 512 token context) is evaluated as an ablation; backbone comparison, adapter configurations, and DeBERTa-specific results are in \S\ref{app:models} and \S\ref{app:ablations}.
The heterogeneous LoRA configuration follows \cref{tab:ensemble_arch} in \S\ref{app:implementation} with $K{=}5$ adapters; LoRA and energy-head dropout are both 0.2.

\vspace{-2pt}
\textbf{Training.}
Each scorer is trained per-dataset via the Bradley--Terry loss in \cref{eq:bt_loss}; candidates are labelled by task-specific verification (answer matching for GSM8K and MuSR, constraint-violation scoring for TravelPlanner, execution-based testing for TACO, and exact-match with statement-consistency checking for K\&K).
Optimiser: AdamW~\citep{loshchilov_2019_adamw} with cosine annealing~\citep{loshchilov_2017_sgdr}; learning rates $5 \times 10^{-5}$ (K\&K $2 \times 10^{-5}$); per-dataset batch sizes and stopping criteria in \cref{tab:training_config}; all training on 3 NVIDIA L40S (46\,GB).

\vspace{-2pt}
\textbf{Inference.}
For the energy decomposition (\cref{eq:energy}), we set $\lambda{=}1.0$ for GSM8K, MuSR, and TACO (where $\Econstraint$ is not applied) and for K\&K (where the exact statement-consistency checker requires no upweighting); $\lambda{=}2.0$ for TravelPlanner, where mean violation is the primary metric and heavier constraint weighting improves it. $\lambda$ values were selected on a held-out validation split.
The $\sigma$-triage rule (\cref{eq:triage}) uses thresholds $\theta_\sigma{=}0.8$ and $\theta_{\text{abstain}}{=}1.5$, both chosen by grid search on a held-out validation split to maximise the pass@1 lift from regeneration while keeping the abstention rate below 10\%.
All evaluations use candidate shuffling (seed 42) to eliminate ordering bias.

\vspace{-8pt}
\subsection{Results}
\label{sec:exp:results}
\vspace{-5pt}

Our 149M Distributional EBM is the strongest non-oracle method on TravelPlanner, within 0.3\,pp 

of the best non-oracle on GSM8K, 4.3\,pp on MuSR, near-greedy on TACO, and within 0.1\,pp of perfect on K\&K (\cref{tab:main_results}), one to three orders of magnitude smaller than every other learned baseline. We organise the analysis around the five experimental questions: \textbf{(i)}~selection vs.\ generation, \textbf{(ii)}~decomposition, \textbf{(iii)}~$\sigma$-triage, \textbf{(iv)}~confounding, and \textbf{(v)}~cross-domain transfer results are in \cref{tab:transfer}.

\vspace{-15pt}
\begin{table}[!h]
\centering
\setlength{\tabcolsep}{4pt}
\caption{Selection accuracy across five benchmarks. \colorbox{gray!15}{Shaded}: our 149M-parameter scorer reranking 32 candidates from a four-generator 7B-class pool; row labels list each baseline's parameter count ($\sim$frontier $\approx$ several hundred B). Frontier models are single-shot at $T{=}0$. \textbf{Bold} = best non-oracle, \underline{underlined} = second-best. mean viol.\ = mean constraint violation (lower better, primary TravelPlanner metric).}
\label{tab:main_results}
\resizebox{\textwidth}{!}{%
\begin{tabular}{@{}lccccc@{}}
    \toprule
    Strategy & GSM8K & MuSR & TravelPlanner$^\ddagger$ & TACO & K\&K \\
    & (pass@1 $\uparrow$) & (pass@1 $\uparrow$) & (mean viol.\ $\downarrow$) & (pass@1 $\uparrow$) & (pass@1 $\uparrow$) \\
    \midrule
    Greedy ($T{=}0$) & 91.8\% & 61.2\% & 0.098 & 86.4\% & \textbf{100.0\%} \\
    Random & 92.7\% & 56.3\% & 0.231 & 46.6\% & 58.7\% \\
    Self-consistency & \textbf{97.3\%} & 64.4\% & \underline{0.096} & 36.4\% & 56.4\% \\
    \midrule
    EORM~\citep{jiang_2025_eorm} (55M)$^\sharp$ & 90.7\% & -- & -- & -- & -- \\
    Skywork-Reward (8B)~\citep{liu_2024_skywork} & 82.1\% & 62.3\% & 0.114$^\ast$ & 70.5\% & 96.4\% \\
    Math-Shepherd PRM (7B)~\citep{wang_2024_mathshepherd} & 94.2\% & -- & -- & -- & -- \\
    LLM-as-Judge ($\sim$frontier, 32 calls/problem)~\citep{zheng_2023_llmjudge} & 99.2\%$^\flat$ & \textbf{73.5\%} & 0.084$^\ast$ & 73.9\% & \underline{99.9\%} \\
    \midrule
    Claude Opus 4.6 ($\sim$frontier, single-shot) & \underline{97.0\%} & \underline{72.0\%} & 0.098 & \textbf{100.0\%} & 99.7\% \\
    Claude Sonnet 4.6 ($\sim$frontier, single-shot) & \underline{97.0\%} & 68.0\% & 0.103 & \underline{95.5\%} & 98.7\% \\
    Qwen-72B-Instruct (72B) & 95.4\% & 54.9\% & 0.121 & 64.8\% & 39.0\% \\
    \midrule
    Oracle & 98.5\% & 86.1\% & 0.028 & 100.0\% & 100.0\% \\
    \midrule
    \rowcolor{gray!15} \textbf{Distributional EBM (Ours, 149M)} & \underline{97.0\%} & 67.7\% & \textbf{0.046} & 88.6\% & \underline{99.9\%} \\
    \bottomrule
\end{tabular}%
}
\vspace{-12pt}
\end{table}

Across the comparisons as shown in \cref{tab:main_results}, Distributional EBM is the strongest non-oracle method on TravelPlanner, within 0.3\,pp of the best non-oracle on GSM8K, within 4.3\,pp on MuSR, near-greedy on TACO, and within 0.1\,pp of perfect on K\&K, while being one to three orders of magnitude smaller than every other learned baseline. \emph{Notes:} $^\ddagger$TP pass@1 (zero-violation) compresses methods to a 0--30\% band: ours 28.0\%, random 14.0\%, self-consistency 12.0\%. $^\ast$Skywork and LLM-as-Judge TP violations are on a 20-problem subset matched to ground-truth scores (full pass@1: 21.1\%, 55.0\%). $^\sharp$EORM uses 256 candidates from 5 LLMs ($8\times$ our budget). $^\flat$LLM-as-Judge GSM8K is on a more lenient pool, see \S\ref{app:datasets}. \texttt{--}: baseline does not apply (math-only PRM/EORM).

\begin{figure}[!ht]
\centering
\begin{minipage}[t]{0.48\textwidth}
\centering
\includegraphics[width=0.85\linewidth]{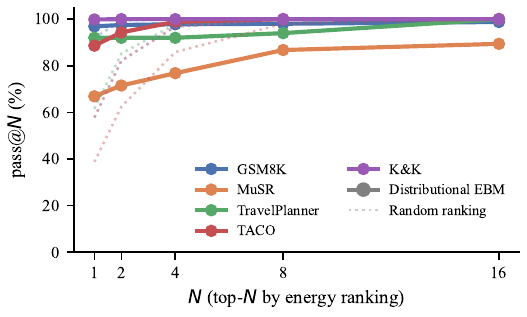}
\vspace{-3pt}
\caption{Pass@$N$~\citep{chen_2021_codex} curves. The Distributional EBM front-loads correct candidates: pass@2 reaches 97.5\% on GSM8K and 71.5\% on MuSR. TravelPlanner uses a relaxed violation threshold ($\theta{=}0.15$).}
\label{fig:pass_at_n}
\end{minipage}%
\hfill
\begin{minipage}[t]{0.49\textwidth}
\centering
\includegraphics[width=0.85\linewidth]{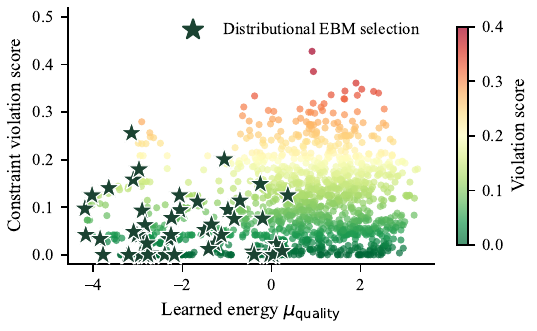}
\vspace{-3pt}
\caption{Energy decomposition on TravelPlanner ($n{=}1{,}487$ parseable candidates; both axes: lower $\downarrow$ better). Stars mark Distributional EBM selections. $\muq$ discriminates horizontally; $\Econstraint$ vertically. Selections cluster in the low-energy, low-violation corner.}
\label{fig:decomposition_scatter}
\end{minipage}
\vspace{-15pt}
\end{figure}

\textbf{(i) Selection vs.\ generation: before vs.\ after the verifier.}
\label{sec:exp:q1}
On the same 32-candidate pool (\cref{tab:main_results}):
On \textbf{MuSR}, greedy 61.2\% / oracle@32 86.1\%; the Distributional EBM closes 26\% of the gap to 67.7\%, exceeding self-consistency (64.4\%) and the $54\times$ larger 8B Skywork-Reward~\citep{liu_2024_skywork}. LLM-as-Judge~\citep{zheng_2023_llmjudge} (Sonnet~4.6) reaches 73.5\% (+5.8\,pp) at 32 frontier-API calls per problem versus our single forward pass: roughly two orders of magnitude cheaper per accuracy point.
On \textbf{TACO}, greedy reaches 86.4\% (oracle 100.0\%); the EBM lifts pass@1 to 88.6\% ($+2.2$\,pp over greedy, $+42.0$\,pp over random, $+52.2$\,pp over self-consistency whose majority vote on extracted code outputs is essentially uninformative), beating Skywork-Reward by $+18.1$\,pp.
On \textbf{K\&K}, greedy already saturates from the strongest single generator (Gemma-4-26B at 100.0\%); the multi-pool EBM achieves 99.9\% (699/700) and dominates Skywork-Reward by $+32$\,pp on 2-character puzzles to $+47$\,pp on 8-character puzzles (\cref{tab:kk_difficulty}).
On \textbf{GSM8K}, contamination compresses the gap (greedy 91.8\%, oracle 98.5\%); the EBM reaches 97.0\%, matching self-consistency within 0.3\,pp and surpassing the 7B Math-Shepherd PRM~\citep{wang_2024_mathshepherd} (94.2\% with explicit step-level supervision) and EORM~\citep{jiang_2025_eorm} (90.7\%; both restricted to math reasoning by construction).
On \textbf{TravelPlanner}, where mean violation is the headline metric: greedy 0.098, EBM 0.046, a 53\% reduction over greedy and 80\% over random.
The lift exists because the multi-generator pool has oracle headroom to convert: a Qwen-only pool collapses the ceiling on the uncontaminated reasoning tasks ($+13.3$/$+52.2$/$+59.8$\,pp on MuSR/TACO/K\&K; \cref{tab:single_multi}). Heterogeneous generators produce diverse error patterns and a higher oracle ceiling, giving the verifier more learning signal and more correct candidates to rank.
\Cref{fig:pass_at_n} shows how the Distributional EBM front-loads correct candidates in its energy ranking; energy separation metrics are in \S\ref{app:energy}.

\textbf{(ii) Learned quality + analytical constraints.}
\label{sec:exp:q2}
On TravelPlanner, the combined energy of \cref{eq:energy} reaches 0.046 mean violation (\cref{tab:main_results}), an 80\% reduction over random (0.231); Skywork-Reward, lacking any structural verification, satisfies all constraints on only 21\% of problems versus our method's 28\% pass@1.
Ablating the two terms of \cref{eq:energy} reveals complementary contributions: $\Equality$ alone achieves 0.058 mean violation, $\Econstraint$ alone achieves 0.038, and the combination (0.046) outperforms $\Equality$ alone while adding learned quality dimensions that the constraint checker does not cover.
\Cref{fig:decomposition_scatter} visualises the two-dimensional energy landscape; the two components agree on the top candidate in only 2\% of problems, and in 28\% of cases the combined energy selects a candidate that neither component would choose independently.
On tasks without analytical constraints (GSM8K, MuSR, TACO), the method reduces to distributional $\muq$ reranking alone and still provides substantial gains.
TravelPlanner improvements over frontier models stem from the analytical constraint checker catching hallucinated transport routes (24-28\% satisfaction for frontier models) that LLMs cannot self-detect. Independent LLM judges (Opus~4.6, Sonnet~4.6) corroborate the violation score and rate EBM-selected plans above oracle picks; full statistics in \S\ref{app:travelplanner_frontier}.
Where structural verification is available, the verifier captures signal frontier models cannot self-detect; where pretraining-scale priors dominate (narrative, code semantics), frontier models lead: the two routes are complementary, not competing. Per-difficulty breakdowns for TACO and K\&K are in \S\ref{app:difficulty}.

\begin{figure}[!h]
    \centering
    \includegraphics[width=\textwidth]{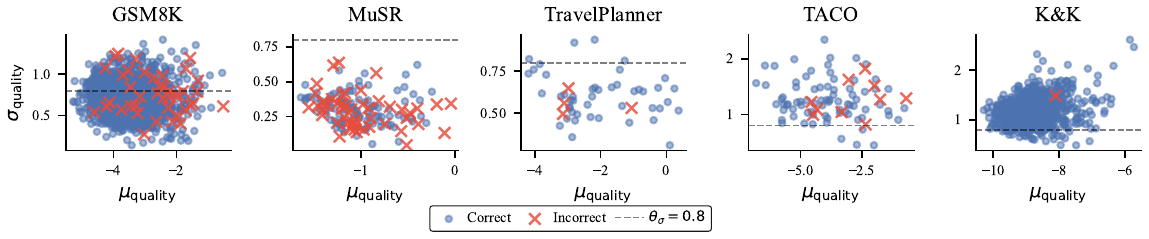}
    \vspace{-16pt}
    \caption{$\sigma$-triage decision surface ($\muq$ vs.\ $\sigmaq$ per problem). On GSM8K, incorrect selections (red) are interspersed with correct ones. On MuSR, all $\sigmaq < \theta_\sigma{=}0.8$ (dashed), so triage never triggers. On TravelPlanner, the wider $\sigmaq$ spread reflects continuous violation labels.}
    \label{fig:triage_surface}
    \vspace{-5pt}
\end{figure}

\textbf{(iii) Ensemble uncertainty as a decision signal.}
\label{sec:exp:q3}
Two-pass $\sigma$-triage (\cref{eq:triage}) helps when pass-1 has headroom and feedback is structured: on TravelPlanner (Qwen-only generator pool, used to isolate the feedback effect from candidate diversity), it improves pass@1 from 74.0\% to 80.0\% (+6.0\,pp).
On TACO (deconfounded), two-pass is neutral (88.6\%); on GSM8K, it is mildly counterproductive ($-1.7$\,pp) precisely because the generator has memorised the answer and feedback distracts retrieval, a diagnostic feature: $\sigma$-triage detects which regime each task falls into rather than applying a one-size-fits-all loop.
For selective prediction~\citep{geifman_2017_selective}, $\sigmaq$ is task-dependent: on TravelPlanner, abstaining on 40\% of problems raises pass@1 from 28.0\% to 36.7\% (+8.7\,pp); on MuSR, $\sigma$ does not signal correctness (AUROC 0.44), and the verifier is honest about where it cannot triage rather than offering an unreliable signal. The when-it-helps regime appears to depend on constraint-checkability and label discreteness rather than scoring task; we discuss the pattern in \S\ref{app:twopass}.
\Cref{fig:triage_surface} visualises the $(\muq, \sigmaq)$ space; full results are in \S\ref{app:twopass} and \S\ref{app:selective}.

\vspace{-5pt}

\textbf{(iv) Reasoning quality vs.\ model-identity shortcuts.}
\label{sec:exp:q4}
When candidate generators have different accuracy levels, the scorer may learn to identify which model produced each candidate rather than evaluating reasoning quality~\citep{geirhos_2020_shortcut}. A cross-dataset confounding analysis (per-model pick distributions and cross-model correct-energy spreads) finds no fingerprinting on four of five tasks: GSM8K, MuSR, TravelPlanner, and K\&K all show near-uniform picks and small spreads despite large generator accuracy gaps (full numbers in \S\ref{app:confounding}). On TACO the initial scorer fingerprinted strongly; a causal style-swap test confirms a formatting shortcut, mitigated via deep feature reweighting (DFR;~\citealp{kirichenko_2023_dfr}) on the energy heads only (0.4\% of parameters), restoring uniform picks at 88.6\% pass@1 with $+13.3$\,pp on HARD and $+14.4$\,pp on UNKNOWN problems.
\textbf{\emph{The determining factor is output format, not generator accuracy gap}}: K\&K has a more extreme accuracy split (88\,pp) than TACO (74\,pp), yet no fingerprinting, because reasoning text is stylistically similar across generators while code contains strong model-specific structural patterns.

\begin{wraptable}{r}{0.55\textwidth}
    \caption{Zero-shot transfer (Distributional EBM pass@1; \colorbox{gray!15}{shaded} = in-domain; \textbf{bold} = best per column; \underline{underlined} = second-best). The TP column reports pass@1 (zero-violation rate) for all rows except the K\&K source row marked $^\diamond$, which reports mean violation (lower better; random $=$ 0.231).}
    \label{tab:transfer}
    \centering
    \small
    \resizebox{\linewidth}{!}{%
    \begin{tabular}{@{}lccccc@{}}
        \toprule
        Source & GSM8K & MuSR & TP & TACO & K\&K \\
        \midrule
        GSM8K & \cellcolor{gray!15}\textbf{97.0} & \underline{64.9} & 0.0 & 39.8 & \underline{92.4} \\
        MuSR & 93.9 & \cellcolor{gray!15}\textbf{67.7} & 0.0 & 36.4 & 89.0 \\
        TP & 93.1 & 60.9 & \cellcolor{gray!15}\underline{28.0} & \underline{56.4} & 33.5 \\
        TACO & 92.5 & 58.9 & \textbf{43.3} & \cellcolor{gray!15}\textbf{88.6} & 31.1 \\
        K\&K & \underline{94.9} & \underline{64.9} & 0.1$^\diamond$ & 51.1 & \cellcolor{gray!15}\textbf{99.9} \\
        \midrule
        SC & 96.1 & 64.4 & 12.0 & 36.4 & 56.4 \\
        Rand & 91.7 & 53.0 & 14.0 & 46.6 & 58.7 \\
        \bottomrule
    \end{tabular}%
    }
    \vspace{-8pt}
\end{wraptable}
\textbf{(v) Cross-domain transfer.}
\label{sec:exp:q5}
To test whether the quality scorer captures transferable reasoning signals, we evaluate each dataset's checkpoint on every other dataset zero-shot (\textit{i.e.}, no retraining).
\Cref{tab:transfer} reports the key results.
Three patterns emerge, all explained by training-data difficulty and signal-regime alignment.
First, \textbf{\emph{hard-to-easy transfer succeeds}}: a MuSR-trained scorer achieves 93.9\% on GSM8K zero-shot (96.1\% with the DeBERTa backbone variant; \S\ref{app:transfer}), within 3.4\,pp of in-domain self-consistency (97.3\%) without ever seeing a math problem. High candidate-accuracy entropy (MuSR 57\%) forces the scorer to learn genuine reasoning-quality signals; high contamination (GSM8K 95\% candidate accuracy) leaves few informative negatives, so the GSM8K-trained scorer reaches only 64.9\% on MuSR -- close to self-consistency (64.4\%) but not exceeding it.
Second, reasoning text and code are different signal regimes: reasoning-trained scorers (GSM8K, MuSR) do not transfer to TACO, but TACO$\rightarrow$TravelPlanner transfers positively because code-execution labels capture quality signals that partially generalise to constrained planning (\cref{tab:transfer}).
Beyond cross-domain transfer, the verifier also transfers across \emph{generator distributions}: a Qwen-only-trained scorer evaluated on a multi-model candidate pool reaches 69.5\% pass@1 on MuSR (matching the multi-trained scorer's 67.7\%) and 90.5\% on GSM8K, confirming the quality signal is not coupled to the training-pool generator style (\S\ref{app:transfer:cross_pool}).
Calibration also transfers more broadly than ranking: a GSM8K-trained scorer achieves $\sigma$-AUROC of 0.79 on TravelPlanner zero-shot despite providing no ranking lift, suggesting ensemble disagreement captures a domain-general difficulty signal.
The full transfer matrix including MATH-500 and a heatmap visualisation (\cref{fig:transfer_heatmap}) are in \S\ref{app:transfer}.

\textbf{Limitations.}
\emph{(i) Generator ceiling}: when no correct candidate exists, even perfect scoring cannot succeed (14\% of MuSR; 58\% of TravelPlanner at strict zero-violation). \emph{(ii) Format dependence}: a GSM8K scorer achieves 65.6\% on MATH-500~\citep{hendrycks_2021_math}, \emph{below} random (68.4\%), because it learned format signals rather than reasoning quality. \emph{(iii) Calibration limits}: on MuSR, $\sigma$-AUROC is anti-correlated (0.44), making selective prediction counterproductive. \emph{(iv) Manual constraint design}: $\Econstraint$ must be authored per task; scaling directions in \S\ref{app:impact}. \emph{(v) Single-round feedback}: the two-pass loop adds only one regeneration round and is therefore less powerful than iterative refinement~\citep{lee_2025_mind_evolution, madaan_2023_selfrefine}. Full ablations are in \S\ref{app:ablations}.

\vspace{-5pt}
\section{Conclusion}
\label{sec:conclusion}
\vspace{-3pt}

Across the five reasoning tasks, three lessons emerge: \emph{(i)} a 149M-parameter verifier orchestrating cheap 7--26B open generators outperforms single-shot Qwen-72B on every benchmark and matches frontier Claude models on constraint-heavy planning, demonstrating that the bottleneck in structured reasoning is selection, not parameter scale; \emph{(ii)} the two routes are complementary: structural verification wins when constraints are checkable, while pretraining-scale priors win where they are not (web-scale corpora encode narrative and code-semantic priors a small scorer cannot match; analytical checking exposes hallucinations frontier models cannot self-detect); \emph{(iii)} honest confounding analysis confirms the scorer learns content on four of five tasks, with the lone code-formatting shortcut on TACO fixable via last-layer retraining at no accuracy cost. The transferability of quality signals (a MuSR scorer reaching 93.9\% on GSM8K zero-shot) suggests structured reasoning shares a common quality signature when training data is difficult enough to force genuine discrimination. Three directions follow: replacing the heterogeneous LoRA ensemble with epistemic deep learning for an axiomatically grounded energy in a single forward pass; extending single-round feedback to iterative energy-guided refinement; and richer structured feedback for tasks where two-pass currently fails on opaque signals (\textit{e.g.}, code execution errors).

\section{Acknowledgments}

We are grateful to Adam Crow and Harrison South for their assistance with compute infrastructure and OpenRouter access, and to Tom Godding, Anup Jadhav, Edward Jackson, and Shefali Sharma for many valuable technical discussions and their constructive feedback on the manuscript.

\bibliography{biblio}

@article{carbone_2024_ebm_hitchhiker,
  title     = {A Hitchhiker's Guide to the Relation of Energy-Based Models with Other Generative Models, Sampling and Statistical Physics},
  author    = {Carbone, Davide},
  journal   = {Transactions on Machine Learning Research},
  year      = {2024},
  note      = {arXiv:2406.13661},
}

@article{blondel_2025_arms_ebms,
  title     = {Autoregressive Language Models Are Secretly Energy-Based Models: Insights into the Lookahead Capabilities of Next-Token Prediction},
  author    = {Blondel, Mathieu and Sander, Michael E and Vivier-Ardisson, Germain and Liu, Tianlin and Roulet, Vincent},
  year      = {2025},
  journal   = {arxiv:2512.15605},
}

@article{tan_2025_rl_ebm,
  title={A Theoretical Lens for RL-Tuned Language Models via Energy-Based Models},
  author={Tan, Zhiquan and Hong, Yinrong},
  journal={arxiv:2512.18730},
  year={2025}
}

@article{jiang_2025_eorm,
  title     = {Learning to {R}ank {C}hain-of-{T}hought: {U}sing a {S}mall {M}odel},
  author    = {Jiang, Eric Hanchen and others},
  year      = {2025},
  journal   = {arXiv:2505.14999},
}

@article{chen_2025_ebmcot,
  title     = {Think {C}onsistently, {R}eason {E}fficiently: {E}nergy-{B}ased {C}alibration for {I}mplicit {C}hain-of-{T}hought},
  author    = {Chen, Zhikang and Cui, Sen and Ye, Deheng and Zhang, Yu and Bian, Yatao and Zhu, Tingting},
  year      = {2025},
  journal   = {arxiv:2511.07124},
}

@article{wei_2022_cot,
  title={Chain-of-thought prompting elicits reasoning in large language models},
  author={Wei, Jason and Wang, Xuezhi and Schuurmans, Dale and Bosma, Maarten and Xia, Fei and Chi, Ed and Le, Quoc V and Zhou, Denny and others},
  journal={Advances in neural information processing systems},
  volume={35},
  pages={24824--24837},
  year={2022}
}

@inproceedings{
wang_2023_selfconsistency,
title={Self-Consistency Improves Chain of Thought Reasoning in Language Models},
author={Xuezhi Wang and Jason Wei and Dale Schuurmans and Quoc V Le and Ed H. Chi and Sharan Narang and Aakanksha Chowdhery and Denny Zhou},
booktitle={The Eleventh International Conference on Learning Representations },
year={2023},
url={https://openreview.net/forum?id=1PL1NIMMrw}
}

@article{lightman_2023_lets_verify,
  title     = {Let's Verify Step by Step},
  author    = {Lightman, Hunter and Kosaraju, Vineet and Burda, Yura and Edwards, Harri and Baker, Bowen and Lee, Teddy and Leike, Jan and Schulman, John and Sutskever, Ilya and Cobbe, Karl},
  year      = {2023},
  journal   = {arxiv:2305.20050},
}

@article{khalifa_2025_thinkprm,
  title     = {Process {R}eward {M}odels {T}hat {T}hink},
  author    = {Khalifa, Muhammad and others},
  year      = {2025},
  journal   = {arXiv:2504.16828},
}

@inproceedings{xie_2024_kk,
  title={On Memorization of Large Language Models in Logical Reasoning},
  author={Chulin Xie and Yangsibo Huang and Chiyuan Zhang and Da Yu and Xinyun Chen and Bill Yuchen Lin and Bo Li and Badih Ghazi and Ravi Kumar},
  booktitle={IJCNLP-AACL},
  year={2024},
  url={https://api.semanticscholar.org/CorpusID:273695832}
}

@inproceedings{xie_2024_travelplanner,
  title={TravelPlanner: A Benchmark for Real-World Planning with Language Agents},
  author={Jian Xie and Kai Zhang and Jiangjie Chen and Tinghui Zhu and Renze Lou and Yuandong Tian and Yanghua Xiao and Yu Su},
  booktitle={International Conference on Machine Learning},
  year={2024},
  url={https://api.semanticscholar.org/CorpusID:267406800}
}

@article{cobbe_2021_gsm8k,
  title={Training verifiers to solve math word problems, 2021},
  author={Cobbe, Karl and Kosaraju, Vineet and Bavarian, Mohammad and Chen, Mark and Jun, Heewoo and Kaiser, Lukasz and Plappert, Matthias and Tworek, Jerry and Hilton, Jacob and Nakano, Reiichiro and others},
  journal={URL https://arxiv.org/abs/2110.14168},
  volume={9},
  year={2021}
}

@article{lee_2025_mind_evolution,
  title     = {Evolving Deeper {LLM} Thinking},
  author    = {Lee, Kuang-Huei and Fischer, Ian and Wu, Yueh-Hua and Marwood, Dave and Baluja, Shumeet and Schuurmans, Dale and Chen, Xinyun},
  year      = {2025},
  journal   = {arxiv:2501.09891},
}

@inproceedings{
snell_2024_scaling_testtime,
title={Scaling {LLM} Test-Time Compute Optimally Can be More Effective than Scaling Parameters for Reasoning},
author={Charlie Victor Snell and Jaehoon Lee and Kelvin Xu and Aviral Kumar},
booktitle={The Thirteenth International Conference on Learning Representations},
year={2025},
url={https://openreview.net/forum?id=4FWAwZtd2n}
}

@article{brown_2024_monkeys,
  title     = {Large Language Monkeys: Scaling Inference Compute with Repeated Sampling},
  author    = {Brown, Bradley and Juravsky, Jordan and Ehrlich, Ryan and Clark, Ronald and Le, Quoc V and R{\'e}, Christopher and Mirhoseini, Azalia},
  year      = {2024},
  journal   = {arxiv:2407.21787},
}

@inproceedings{xia_2025_agentrm,
    title = "{A}gent{RM}: Enhancing Agent Generalization with Reward Modeling",
    author = "Xia, Yu  and
      Fan, Jingru  and
      Chen, Weize  and
      Yan, Siyu  and
      Cong, Xin  and
      Zhang, Zhong  and
      Lu, Yaxi  and
      Lin, Yankai  and
      Liu, Zhiyuan  and
      Sun, Maosong",
    editor = "Che, Wanxiang  and
      Nabende, Joyce  and
      Shutova, Ekaterina  and
      Pilehvar, Mohammad Taher",
    booktitle = "Proceedings of the 63rd Annual Meeting of the Association for Computational Linguistics (Volume 1: Long Papers)",
    month = jul,
    year = "2025",
    address = "Vienna, Austria",
    publisher = "Association for Computational Linguistics",
    url = "https://aclanthology.org/2025.acl-long.945/",
    doi = "10.18653/v1/2025.acl-long.945",
    pages = "19277--19290",
    ISBN = "979-8-89176-251-0"
}

@article{gleave_2022_uncertainty_rm,
  title     = {Uncertainty Estimation for Language Reward Models},
  author    = {Gleave, Adam and Irving, Geoffrey},
  year      = {2022},
  journal   = {arXiv:2203.07472},
}

@article{fort_2019_deep_ensembles,
  title     = {Deep Ensembles: A Loss Landscape Perspective},
  author    = {Fort, Stanislav and Hu, Huiyi and Lakshminarayanan, Balaji},
  year      = {2019},
  journal   = {arXiv:1912.02757},
}

@inproceedings{lakshminarayanan_2017_simple_ensembles,
  title     = {Simple and {S}calable {P}redictive {U}ncertainty {E}stimation {U}sing {D}eep {E}nsembles},
  author    = {Lakshminarayanan, Balaji and Pritzel, Alexander and Blundell, Charles},
  booktitle = {Advances in Neural Information Processing Systems (NeurIPS)},
  year      = {2017},
  url={https://arxiv.org/abs/1612.01474},
  
}

@misc{he_2023_debertav3,
      title={DeBERTaV3: Improving {D}e{BERT}a using {ELECTRA}-{S}tyle {P}re-{T}raining with {G}radient-{D}isentangled {E}mbedding {S}haring}, 
      author={Pengcheng He and Jianfeng Gao and Weizhu Chen},
      year={2023},
      eprint={2111.09543},
      archivePrefix={arXiv},
      primaryClass={cs.CL},
      url={https://arxiv.org/abs/2111.09543}, 
}

@article{qwen_2024_qwen25,
  title     = {Qwen2.5 Technical Report},
  author    = {{Qwen Team}},
  year      = {2024},
  journal   = {arXiv:2412.15115},
}

@article{bradley_1952_paired,
  title     = {Rank Analysis of Incomplete Block Designs: {I}. {T}he Method of Paired Comparisons},
  author    = {Bradley, Ralph Allan and Terry, Milton E.},
  journal   = {Biometrika},
  volume    = {39},
  number    = {3/4},
  pages     = {324--345},
  year      = {1952},
  publisher={JSTOR}
}

@article{brown_2020_gpt3,
  title={Language models are few-shot learners},
  author={Brown, Tom and Mann, Benjamin and Ryder, Nick and Subbiah, Melanie and Kaplan, Jared D and Dhariwal, Prafulla and Neelakantan, Arvind and Shyam, Pranav and Sastry, Girish and Askell, Amanda and others},
  journal={Advances in neural information processing systems},
  volume={33},
  pages={1877--1901},
  year={2020}
}

@techreport{openai_2023_gpt4,
  title         = {{GPT-4} Technical Report},
  author        = {OpenAI},
  institution   = {OpenAI},
  year          = {2023},
  note          = {arXiv:2303.08774},
}

@inproceedings{
zhou_2023_leasttomost,
title={Least-to-Most Prompting Enables Complex Reasoning in Large Language Models},
author={Denny Zhou and Nathanael Sch{\"a}rli and Le Hou and Jason Wei and Nathan Scales and Xuezhi Wang and Dale Schuurmans and Claire Cui and Olivier Bousquet and Quoc V Le and Ed H. Chi},
booktitle={The Eleventh International Conference on Learning Representations },
year={2023},
url={https://openreview.net/forum?id=WZH7099tgfM}
}

@inproceedings{
yao_2023_tot,
title={Tree of Thoughts: Deliberate Problem Solving with Large Language Models},
author={Shunyu Yao and Dian Yu and Jeffrey Zhao and Izhak Shafran and Thomas L. Griffiths and Yuan Cao and Karthik R Narasimhan},
booktitle={Thirty-seventh Conference on Neural Information Processing Systems},
year={2023},
url={https://openreview.net/forum?id=5Xc1ecxO1h}
}

@inproceedings{sprague_2024_musr,
 author = {Sprague, Zayne and Ye, Xi and Bostrom, Kaj and Chaudhuri, Swarat and Durrett, Greg},
 booktitle = {International Conference on Learning Representations},
 editor = {B. Kim and Y. Yue and S. Chaudhuri and K. Fragkiadaki and M. Khan and Y. Sun},
 pages = {14670--14728},
 title = {MuSR: Testing the Limits of Chain-of-thought with Multistep Soft Reasoning},
 volume = {2024},
 year = {2024}
}

@article{li_2023_taco,
  title     = {{TACO}: Topics in Algorithmic COde generation dataset},
  author    = {Li, Rongao and Fu, Jie and Zhang, Bo-Wen and Huang, Tao and Sun, Zhihong and Lyu, Chen and Liu, Guang and Jin, Zhi and Li, Ge},
  year      = {2023},
  journal   = {arxiv:2312.14852},
}

@techreport{openai_2025_gptoss,
  title       = {Introducing gpt-oss},
  author      = {OpenAI},
  institution = {OpenAI},
  year        = {2025},
  note        = {\url{https://openai.com/index/introducing-gpt-oss/}},
}

@inproceedings{warner_2024_modernbert,
  title={Smarter, better, faster, longer: A modern bidirectional encoder for fast, memory efficient, and long context finetuning and inference},
  author={Warner, Benjamin and Chaffin, Antoine and Clavi{\'e}, Benjamin and Weller, Orion and Hallstr{\"o}m, Oskar and Taghadouini, Said and Gallagher, Alexis and Biswas, Raja and Ladhak, Faisal and Aarsen, Tom and others},
  booktitle={Proceedings of the 63rd Annual Meeting of the Association for Computational Linguistics (Volume 1: Long Papers)},
  pages={2526--2547},
  year={2025}
}

@article{grattafiori_2024_llama3,
  title={The llama 3 herd of models},
  author={Grattafiori, Aaron and Dubey, Abhimanyu and Jauhri, Abhinav and Pandey, Abhinav and Kadian, Abhishek and Al-Dahle, Ahmad and Letman, Aiesha and Mathur, Akhil and Schelten, Alan and Vaughan, Alex and others},
  journal={arxiv:2407.21783},
  year={2024}
}

@techreport{google_2026_gemma,
    title       = {Gemma 4 Technical Report},
    author      = {{Google DeepMind}},
    institution = {Google DeepMind},
    year        = {2026},
}

@techreport{anthropic_2026_claude,
    title       = {The Claude Model Family: Claude Opus 4.6, Claude Sonnet 4.6},
    author      = {Anthropic},
    institution = {Anthropic},
    year        = {2026},
    note        = {\url{https://docs.anthropic.com/en/docs/about-claude/models}},
}

@article{geirhos_2020_shortcut,
  title     = {Shortcut Learning in Deep Neural Networks},
  author    = {Geirhos, Robert and Jacobsen, J{\"o}rn-Henrik and Michaelis, Claudio and Zemel, Richard and Brendel, Wieland and Bethge, Matthias and Wichmann, Felix A},
  journal   = {Nature Machine Intelligence},
  volume    = {2},
  number={11},
  pages     = {665--673},
  year      = {2020},
  publisher={Nature Publishing Group UK London}
}

@inproceedings{loshchilov_2019_adamw,
  title     = {Decoupled Weight Decay Regularization},
  author    = {Loshchilov, Ilya and Hutter, Frank},
  booktitle = {International Conference on Learning Representations (ICLR)},
  year      = {2019},
  note      = {arXiv:1711.05101},
}

@inproceedings{ouyang_2022_instructgpt,
  title     = {Training Language Models to Follow Instructions with Human Feedback},
  author    = {Ouyang, Long and Wu, Jeffrey and Jiang, Xu and Almeida, Diogo and Wainwright, Carroll and Mishkin, Pamela and Zhang, Chong and Agarwal, Sandhini and Slama, Katarina and Ray, Alex and others},
  booktitle = {Advances in Neural Information Processing Systems (NeurIPS)},
  year      = {2022},
  note      = {arXiv:2203.02155},
}

@article{uesato_2022_prm,
  title={Solving math word problems with process- and outcome-based feedback},
  author={Jonathan Uesato and Nate Kushman and Ramana Kumar and Francis Song and Noah Siegel and L. Wang and Antonia Creswell and Geoffrey Irving and Irina Higgins},
  journal={ArXiv},
  year={2022},
  volume={abs/2211.14275},
  url={https://api.semanticscholar.org/CorpusID:254017497}
}

@article{deepseek_2025_r1,
  title     = {{DeepSeek-R1}: Incentivizing Reasoning Capability in {LLMs} via Reinforcement Learning},
  author={Guo, Daya and Yang, Dejian and Zhang, Haowei and Song, Junxiao and Wang, Peiyi and Zhu, Qihao and Xu, Runxin and Zhang, Ruoyu and Ma, Shirong and Bi, Xiao and others},
  year      = {2025},
  journal   = {arxiv:2501.12948},
}

@inproceedings{houlsby_2019_adapters,
  title     = {Parameter-Efficient Transfer Learning for {NLP}},
  author    = {Houlsby, Neil and Giurgiu, Andrei and Jastrzebski, Stanislaw and Morrone, Bruna and de Laroussilhe, Quentin and Gesmundo, Andrea and Attariyan, Mona and Gelly, Sylvain},
  booktitle = {International Conference on Machine Learning (ICML)},
  year      = {2019},
  note      = {arXiv:1902.00751},
}

@inproceedings{stiennon_2020_summarize,
author = {Stiennon, Nisan and Ouyang, Long and Wu, Jeff and Ziegler, Daniel M. and Lowe, Ryan and Voss, Chelsea and Radford, Alec and Amodei, Dario and Christiano, Paul},
title = {Learning to summarize from human feedback},
year = {2020},
isbn = {9781713829546},
publisher = {Curran Associates Inc.},
address = {Red Hook, NY, USA},
booktitle = {Proceedings of the 34th International Conference on Neural Information Processing Systems},
articleno = {253},
numpages = {14},
location = {Vancouver, BC, Canada},
series = {NIPS '20}
}

@inproceedings{oren_2024_contamination,
  title     = {Proving Test Set Contamination in Black Box Language Models},
  author    = {Oren, Yonatan and Meister, Nicole and Bhatt, Niladri and Raji, Inioluwa Deborah and Cotterell, Ryan},
  booktitle = {International Conference on Learning Representations (ICLR)},
  year      = {2024},
  note      = {arXiv:2310.17623},
}

@inproceedings{hu_2022_lora,
  title     = {{LoRA}: Low-Rank Adaptation of Large Language Models},
  author    = {Hu, Edward J. and Shen, Yelong and Wallis, Phillip and Allen-Zhu, Zeyuan and Li, Yuanzhi and Wang, Shean and Wang, Lu and Chen, Weizhu},
  booktitle = {International Conference on Learning Representations (ICLR)},
  year      = {2022},
  note      = {arXiv:2106.09685},
}

@misc{hendrycks_2021_math,
      title={Measuring {M}athematical {P}roblem {S}olving {W}ith the {MATH} {D}ataset}, 
      author={Dan Hendrycks and Collin Burns and Saurav Kadavath and Akul Arora and Steven Basart and Eric Tang and Dawn Song and Jacob Steinhardt},
      year={2021},
      eprint={2103.03874},
      archivePrefix={arXiv},
      primaryClass={cs.LG},
      url={https://arxiv.org/abs/2103.03874}, 
}

@inproceedings{kirichenko_2023_dfr,
  title     = {Last {L}ayer {R}e-{T}raining is {S}ufficient for {R}obustness to {S}purious {C}orrelations},
  author    = {Kirichenko, Polina and Izmailov, Pavel and Wilson, Andrew Gordon},
  booktitle = {International Conference on Learning Representations (ICLR)},
  year      = {2023},
  note      = {arXiv:2204.02937},
}

@article{yang_2023_rethink_benchmark,
  title={Rethinking Benchmark and Contamination for Language Models with Rephrased Samples},
  author={Shuo Yang and Wei-Lin Chiang and Lianmin Zheng and Joseph E. Gonzalez and Ion Stoica},
  journal={ArXiv},
  year={2023},
  volume={abs/2311.04850},
  url={https://api.semanticscholar.org/CorpusID:265050721}
}

@inproceedings{kojima_2022_zero_shot_cot,
  title     = {Large {L}anguage {M}odels are {Z}ero-{S}hot {R}easoners},
  author    = {Kojima, Takeshi and Gu, Shixiang Shane and Reid, Machel and Matsuo, Yutaka and Iwasawa, Yusuke},
  booktitle = {Advances in Neural Information Processing Systems (NeurIPS)},
  year      = {2022},
  note      = {arXiv:2205.11916},
}

@inproceedings{christiano_2017_rlhf,
  author = {Christiano, Paul F and Leike, Jan and Brown, Tom and Martic, Miljan and Legg, Shane and Amodei, Dario},
 booktitle = {Advances in Neural Information Processing Systems},
 editor = {I. Guyon and U. Von Luxburg and S. Bengio and H. Wallach and R. Fergus and S. Vishwanathan and R. Garnett},
 pages = {},
 publisher = {Curran Associates, Inc.},
 title = {Deep Reinforcement Learning from Human Preferences},
 url = {https://proceedings.neurips.cc/paper_files/paper/2017/file/d5e2c0adad503c91f91df240d0cd4e49-Paper.pdf},
 volume = {30},
 year = {2017}
}

@inproceedings{rafailov_2023_dpo,
  title     = {Direct Preference Optimization: Your Language Model is Secretly a Reward Model},
  author    = {Rafailov, Rafael and Sharma, Archit and Mitchell, Eric and Ermon, Stefano and Manning, Christopher D. and Finn, Chelsea},
  booktitle = {Advances in Neural Information Processing Systems (NeurIPS)},
  year      = {2023},
  note      = {arXiv:2305.18290},
}

@inproceedings{gal_2016_dropout,
  title     = {Dropout as a {Bayesian} Approximation: Representing Model Uncertainty in Deep Learning},
  author={Gal, Yarin and Ghahramani, Zoubin},
  booktitle={international conference on machine learning},
  pages={1050--1059},
  year={2016},
  organization={PMLR}
}

@inproceedings{madaan_2023_selfrefine,
  title     = {Self-Refine: Iterative Refinement with Self-Feedback},
  author    = {Madaan, Aman and Tandon, Niket and Gupta, Prakhar and Hallinan, Skyler and Gao, Luyu and Wiegreffe, Sarah and Alon, Uri and Dziri, Nouha and Prabhumoye, Shrimai and Yang, Yiming and Gupta, Shashank and Majumder, Bodhisattwa Prasad and Hermann, Katherine M. and Welleck, Sean and Yazdanbakhsh, Amir and Clark, Peter},
  booktitle = {Advances in Neural Information Processing Systems (NeurIPS)},
  year      = {2023},
  note      = {arXiv:2303.17651},
}

@inproceedings{deng_2020_residual_ebm,
  title={Residual energy-based models for text generation},
  author={Deng, Yuntian and Bakhtin, Anton and Ott, Myle and Szlam, Arthur and Ranzato, Marc'Aurelio},
  booktitle={International Conference on Learning Representations (ICLR)},
  year={2020},
  note={arXiv:2004.11714}
}

@inproceedings{
wang_2024_moa,
title={Mixture-of-Agents Enhances Large Language Model Capabilities},
author={Junlin Wang and Jue WANG and Ben Athiwaratkun and Ce Zhang and James Zou},
booktitle={The Thirteenth International Conference on Learning Representations},
year={2025},
url={https://openreview.net/forum?id=h0ZfDIrj7T}
}

@inproceedings{devlin_2019_bert,
  title     = {{BERT}: Pre-training of Deep Bidirectional Transformers for Language Understanding},
  author    = {Devlin, Jacob and Chang, Ming-Wei and Lee, Kenton and Toutanova, Kristina},
  booktitle = {Proceedings of the 2019 Conference of the North American Chapter of the Association for Computational Linguistics (NAACL)},
  pages     = {4171--4186},
  year      = {2019},
  note      = {arXiv:1810.04805}
}

@article{hendrycks_2016_gelu,
  title={Gaussian Error Linear Units (GELUs)},
  author={Dan Hendrycks and Kevin Gimpel},
  journal={arXiv: Learning},
  year={2016},
  url={https://api.semanticscholar.org/CorpusID:125617073}
}

@article{ba_2016_layernorm,
  title={Layer Normalization},
  author={Jimmy Ba and Jamie Ryan Kiros and Geoffrey E. Hinton},
  journal={ArXiv},
  year={2016},
  volume={abs/1607.06450},
  url={https://api.semanticscholar.org/CorpusID:8236317}
}

@inproceedings{loshchilov_2017_sgdr,
  title     = {{SGDR}: Stochastic Gradient Descent with Warm Restarts},
  author    = {Loshchilov, Ilya and Hutter, Frank},
  booktitle = {International Conference on Learning Representations (ICLR)},
  year      = {2017},
  note      = {arXiv:1608.03983},
}

@inproceedings{holtzman_2020_nucleus,
  title     = {The {C}urious {C}ase of {N}eural {T}ext {D}egeneration},
  author    = {Holtzman, Ari and Buys, Jan and Du, Li and Forbes, Maxwell and Choi, Yejin},
  booktitle = {International Conference on Learning Representations (ICLR)},
  year      = {2020},
  note      = {arXiv:1904.09751},
}

@inproceedings{guo_2017_calibration,
  title={On calibration of modern neural networks},
  author={Guo, Chuan and Pleiss, Geoff and Sun, Yu and Weinberger, Kilian Q},
  booktitle={International conference on machine learning},
  pages={1321--1330},
  year={2017},
  organization={PMLR}
}

@article{geifman_2017_selective,
  title={Shortcut learning in deep neural networks},
  author={Geirhos, Robert and Jacobsen, J{\"o}rn-Henrik and Michaelis, Claudio and Zemel, Richard and Brendel, Wieland and Bethge, Matthias and Wichmann, Felix A},
  journal={Nature Machine Intelligence},
  volume={2},
  number={11},
  pages={665--673},
  year={2020},
  publisher={Nature Publishing Group UK London}
}

@article{chen_2021_codex,
  title     = {Evaluating Large Language Models Trained on Code},
  author={Chen, Mark and Tworek, Jerry and Jun, Heewoo and Yuan, Qiming and Pinto, Henrique Ponde De Oliveira and Kaplan, Jared and Edwards, Harri and Burda, Yuri and Joseph, Nicholas and Brockman, Greg and others},
  journal={arxiv:2107.03374},
  year={2021}
}

@article{breiman_1996_bagging,
  title={Bagging predictors},
  author={Breiman, Leo},
  journal={Machine learning},
  volume={24},
  number={2},
  pages={123--140},
  year={1996},
  publisher={Springer}
}

@article{lecun_2006_ebm_tutorial,
  title     = {A {T}utorial on {E}nergy-{B}ased {L}earning},
  author    = {LeCun, Yann and Chopra, Sumit and Hadsell, Raia and Ranzato, Marc'Aurelio and Huang, Fu Jie},
  journal   = {Predicting Structured Data},
  year      = {2006},
  publisher = {MIT Press},
}

@article{liu_2024_skywork,
  title={Skywork-Reward: Bag of Tricks for Reward Modeling in {LLM}s},
  author={Liu, Chris Yuhao and Zeng, Liang and Liu, Jiacai and Yan, Rui and He, Jujie and Wang, Chaojie and Yan, Shuicheng and Liu, Yang and Zhou, Yahui},
  journal={arXiv preprint arXiv:2410.18451},
  year={2024},
}

@inproceedings{wang_2024_mathshepherd,
  title={Math-Shepherd: Verify and Reinforce {LLM}s Step-by-Step without Human Annotations},
  author={Wang, Peiyi and Li, Lei and Shao, Zhihong and Xu, Runxin and Dai, Damai and Li, Yifei and Chen, Deli and Wu, Yu and Sui, Zhifang},
  booktitle={Proceedings of the 62nd Annual Meeting of the Association for Computational Linguistics (ACL)},
  year={2024},
}

@inproceedings{zheng_2023_llmjudge,
  title={Judging {LLM}-as-a-Judge with {MT}-Bench and Chatbot Arena},
  author={Zheng, Lianmin and Chiang, Wei-Lin and Sheng, Ying and Zhuang, Siyuan and Wu, Zhanghao and Zhuang, Yonghao and Lin, Zi and Li, Zhuohan and Li, Dacheng and Xing, Eric P. and Zhang, Hao and Gonzalez, Joseph E. and Stoica, Ion},
  booktitle={Advances in Neural Information Processing Systems (NeurIPS)},
  year={2023},
}
\bibliographystyle{plainnat}

\newpage
\appendix

\section{Proofs of Theoretical Results}
\label{app:proofs}
\vspace{-5pt}

This section provides full proofs and the detailed consequences/implications of \cref{prop:condorcet} and \cref{prop:variance} from \cref{sec:method:quality}.

\subsection{Proof of \cref{prop:condorcet}: ensemble ranking under correlated adapters}

For each adapter $k$, define the binary correctness indicator
\[
V_k = \mathbf{1}\!\left[E^k_{\text{quality}}(x, y^+) < E^k_{\text{quality}}(x, y^-)\right] \in \{0, 1\},
\]
so that $\mathbb{E}[V_k] = q$ and $\mathrm{Var}(V_k) = q(1-q)$. The exchangeable correlation assumption is $\mathrm{Corr}(V_j, V_k) = \rho$ for all $j \ne k$, equivalently $\mathrm{Cov}(V_j, V_k) = \rho\, q(1-q)$.

Let $V = \sum_{k=1}^{K} V_k$ be the total correct-vote count. Then
\[
\mathbb{E}[V] = K q, \qquad
\mathrm{Var}(V) = K\, q(1-q) + K(K-1)\, \rho\, q(1-q) = K q(1-q)\bigl[1 + (K-1)\rho\bigr].
\]
The ensemble mean $\muq(x, y^+) < \muq(x, y^-)$ iff $V > K/2$, i.e., a strict majority of adapters vote correctly. Under the Gaussian approximation (which the central limit theorem supports for $K \ge 5$ in the parameter range we consider),
\[
V \;\stackrel{\mathrm{approx}}{\sim}\; \mathcal{N}\!\Bigl(K q,\; K q(1-q)\bigl[1 + (K-1)\rho\bigr]\Bigr),
\]
and standardising,
\begin{align*}
P_{\text{ensemble}}(K, q, \rho) &= \mathbb{P}(V > K/2) = \Phi\!\left(\frac{K q - K/2}{\sqrt{K q(1-q)(1+(K-1)\rho)}}\right) \\
&= \Phi\!\left(\frac{\sqrt{K}\,(q - \tfrac{1}{2})}{\sqrt{q(1-q)(1 + (K-1)\rho)}}\right),
\end{align*}
which is \cref{eq:condorcet}.

For the limit, write the argument of $\Phi$ as
\[
\frac{\sqrt{K}\,(q - \tfrac{1}{2})}{\sqrt{q(1-q)(1+(K-1)\rho)}}
=
\frac{(q - \tfrac{1}{2})}{\sqrt{q(1-q)\bigl(\tfrac{1}{K} + \tfrac{K-1}{K}\rho\bigr)}}.
\]
As $K \to \infty$, $\tfrac{1}{K} \to 0$ and $\tfrac{K-1}{K} \to 1$, so for any $\rho > 0$ the argument converges to $(q - \tfrac{1}{2})/\sqrt{q(1-q)\rho}$, giving \cref{eq:condorcet}:
\[
P^\infty(q, \rho) = \Phi\!\left(\frac{q - \tfrac{1}{2}}{\sqrt{q(1-q)\rho}}\right) < 1
\quad \text{for any finite } q \in (\tfrac{1}{2}, 1).
\]
The ceiling is strictly below 1 because the argument of $\Phi$ is finite. Only when $\rho = 0$ (perfectly independent adapters) does $\sqrt{K}$ in the numerator dominate, sending $P_{\text{ensemble}} \to 1$.

\paragraph{Detailed consequences.} Three consequences follow.
\begin{enumerate}[label=(C\arabic*),leftmargin=*]
    \item \textbf{Diversity amplification.} $\partial P_{\text{ensemble}} / \partial \rho < 0$ for $q > \tfrac{1}{2}$: lower inter-adapter correlation strictly improves ensemble accuracy. This is non-obvious because one might expect that highly correlated (i.e., individually strong) adapters would produce a stronger ensemble; the proof shows the opposite.
    \item \textbf{Diminishing returns of $K$ at high correlation.} When $\rho = 1$, all adapters are perfectly correlated and $P_{\text{ensemble}} \to q$ regardless of $K$: adding more identical adapters provides zero benefit. When $\rho < 1$, ensemble accuracy increases with $K$ but saturates at the ceiling $P^\infty(q, \rho)$.
    \item \textbf{Ceiling binding regime.} Empirically, at $K{=}5$, the gap $P^\infty - P_{\text{ensemble}}(K{=}5)$ is small in our regime (less than 1\,pp for typical $(q, \rho)$ pairs we observe), so additional adapters yield strongly diminishing returns once $K \gtrsim 5$. The ceiling is binding for tasks where per-adapter accuracy $q$ is moderate (e.g., narrative reasoning where $q \approx 0.6$) and slack only for tasks where $q$ is near unity (e.g., K\&K with high adapter accuracy).
\end{enumerate}
The implication is that ensemble diversity is necessary but not sufficient: a fundamental ceiling is determined by the joint $(q, \rho)$ of the adapter architecture and training distribution.

\subsection{Proof of \cref{prop:variance}: partitioned variance under energy decomposition}

We have $E(x, y) = \muq(x, y) + \lambda \cdot \Econstraint(x, y)$ where $\Econstraint$ is a deterministic analytical function of $(x, y)$ with no learnable parameters and no stochastic components. Treating $\muq$ as a random variable (over the ensemble's epistemic uncertainty), the variance of a sum of a random and a deterministic quantity is:
\begin{align*}
\mathrm{Var}(E) &= \mathrm{Var}(\muq + \lambda\,\Econstraint) \\
&= \mathrm{Var}(\muq) + \lambda^2\,\underbrace{\mathrm{Var}(\Econstraint)}_{=\,0} + 2\lambda\,\underbrace{\mathrm{Cov}(\muq, \Econstraint)}_{=\,0} = \sigmaq^2,
\end{align*}
since the variance of any deterministic function is $0$ and the covariance of any random variable with a deterministic quantity is $0$. This gives \cref{eq:variance_decomp}.

For the monolithic comparison, suppose a single scorer $E_{\text{mono}}$ is trained jointly to predict both quality and constraint satisfaction, $E_{\text{mono}}(x, y) = \Equality^{\text{mono}}(x, y) + \Econstraint^{\text{mono}}(x, y)$, where now both components are random variables under the scorer's epistemic uncertainty. Then
\[
\mathrm{Var}(E_{\text{mono}}) = \mathrm{Var}(\Equality^{\text{mono}}) + \mathrm{Var}(\Econstraint^{\text{mono}}) + 2\,\mathrm{Cov}(\Equality^{\text{mono}}, \Econstraint^{\text{mono}}).
\]
This equals $\sigmaq^2$ only in the degenerate case where the monolithic scorer learns $\Econstraint^{\text{mono}}$ exactly (so $\mathrm{Var}(\Econstraint^{\text{mono}}) = 0$ and the covariance term vanishes); otherwise the variance mixes constraint-prediction error into the uncertainty signal.

\paragraph{Detailed implications.} Three implications follow.
\begin{enumerate}[label=(I\arabic*),leftmargin=*]
    \item \textbf{Interpretable uncertainty.} $\sigmaq$ measures exactly how uncertain the ensemble is about reasoning quality, with zero contribution from constraint verification. A monolithic scorer's uncertainty conflates both sources and cannot distinguish ``uncertain about quality'' from ``uncertain about constraints.''
    \item \textbf{Strictly lower predictive variance.} The decomposition yields strictly lower predictive variance than a monolithic scorer whenever the analytical constraints have lower error than a learned approximation. This holds by construction for TravelPlanner (database-verified routes) and K\&K (statement-consistency checker), where $\Econstraint$ is exact.
    \item \textbf{Natural decision rule.} The $\sigma$-triage mechanism (\cref{sec:method:twopass}) is the natural decision rule under this partitioned variance:
    \begin{itemize}[leftmargin=*]
        \item \emph{Accept} when both $\sigmaq$ and $\Econstraint$ are low (quality reliable, constraints satisfied).
        \item \emph{Regenerate with constraint feedback} when $\Econstraint$ is high (constraints violated, regardless of $\sigmaq$).
        \item \emph{Regenerate with quality feedback} when $\sigmaq$ is high but $\Econstraint$ is low (constraints fine, quality uncertain).
        \item \emph{Abstain} when $\sigmaq$ exceeds a threshold (quality completely unknown).
    \end{itemize}
\end{enumerate}

\section{Model Details}
\label{app:models}
\vspace{-5pt}

This section details the candidate generators (\cref{tab:generators}), scorer backbones (\cref{tab:backbones}), ensemble architecture (\cref{tab:ensemble_arch}), and training configuration (\cref{tab:training_config}).

\subsection{Candidate Generators}

\begin{table}[h!]
    \caption{Candidate generator models. All models are frozen throughout and accessed via the OpenRouter API. Active parameters are listed for mixture-of-experts models.}
    \label{tab:generators}
    \centering
    \resizebox{\textwidth}{!}{%
    \begin{tabular}{@{}llccl@{}}
        \toprule
        Model & Architecture & Parameters & Context & Role \\
        \midrule
        Qwen-2.5-7B-Instruct~\citep{qwen_2024_qwen25} & Dense Transformer & 7B & 128K & 7B baseline; pass-2 generator \\
        LLaMA-3.1-8B-Instruct~\citep{grattafiori_2024_llama3} & Dense Transformer & 8B & 128K & 8B baseline; highest error diversity \\
        Gemma-4-26B-IT~\citep{google_2026_gemma} & Mixture-of-Experts & 26B (4B active) & 128K & Strongest generator across all tasks \\
        GPT-OSS-20B~\citep{openai_2025_gptoss} & Mixture-of-Experts & 20B (3.6B active) & 128K & Reasoning model (extended thinking) \\
        \bottomrule
    \end{tabular}%
    }
\end{table}

Gemma-4-26B-IT consistently produces the highest candidate accuracy across all four benchmarks (96.7\% on GSM8K, 66.7\% on MuSR, 67.9\% on TravelPlanner, 84.6\% on TACO; see \cref{tab:permodel}).
LLaMA-3.1-8B-Instruct produces the lowest accuracy but contributes the most diverse error patterns: on GSM8K, it provides 1{,}309 unique negative problems, 4-10$\times$ more than the other generators.
GPT-OSS-20B is a reasoning model that outputs extended thinking traces before the final answer; we set \texttt{max\_tokens=4096} to capture both the reasoning and the response.

\subsection{Frontier Models}

Three frontier models are evaluated as single-shot baselines on TravelPlanner and TACO:
Claude Opus~4.6 and Claude Sonnet~4.6~\citep{anthropic_2026_claude}, and Qwen-72B~\citep{qwen_2024_qwen25}.
All generate a single deterministic output at $T{=}0$ with \texttt{max\_tokens=2048}.
Opus~4.6 also serves as the primary LLM judge for the TravelPlanner quality validation (\cref{sec:exp:results}).

\subsection{Scorer Backbones}

\begin{table}[h!]
    \caption{Scorer backbone comparison.}
    \label{tab:backbones}
    \centering
    \begin{tabular}{@{}lccc@{}}
        \toprule
        Backbone & Parameters & Max context & Trainable / backbone \\
        \midrule
        ModernBERT-base~\citep{warner_2024_modernbert} & 149M & 8{,}192 & ${\sim}$3\% \\
        DeBERTa-v3-base~\citep{he_2023_debertav3} & 86M & 512 & ${\sim}$3\% \\
        \bottomrule
    \end{tabular}
\end{table}

ModernBERT-base is the primary backbone for all experiments.
Its 8{,}192-token context accommodates MuSR narratives (3{,}000-6{,}000 characters) and TravelPlanner itineraries without truncation.
DeBERTa-v3-base is evaluated as an ablation; its 512-token context truncates long inputs but produces better-calibrated uncertainty estimates (ECE 0.10 vs.\ 0.32 on GSM8K) and stronger zero-shot transfer (96.1\% on GSM8K from MuSR training vs.\ 93.9\% for ModernBERT).

\section{Implementation Details}
\label{app:implementation}
\vspace{-5pt}

\paragraph{Energy head architecture.} Each adapter $k$ has an independent energy head: $E^k_{\text{quality}}(x, y) = \text{MLP}_k\!\big(\text{LayerNorm}(\mathbf{h}^k_{\text{CLS}})\big)$, where $\text{MLP}_k$ consists of $\text{Linear}(d {\to} d)$, GELU~\citep{hendrycks_2016_gelu}, $\text{Dropout}(p)$, $\text{Linear}(d {\to} 1)$, preceded by LayerNorm~\citep{ba_2016_layernorm}.

\begin{table}[h!]
    \caption{Heterogeneous LoRA ensemble configuration. All adapters share a single frozen encoder backbone. Structural diversity in rank, scaling, and target modules forces disagreement that reflects epistemic uncertainty. ModernBERT targets are shown; DeBERTa uses query, key, value, and output projections analogously.}
    \label{tab:ensemble_arch}
    \centering
    \begin{tabular}{@{}ccccl@{}}
        \toprule
        Adapter & Rank & Alpha & Dropout & Target Modules (ModernBERT) \\
        \midrule
        0 & 8 & 16 & 0.2 & Wqkv \\
        1 & 8 & 16 & 0.2 & Wqkv, Wo \\
        2 & 16 & 32 & 0.2 & Wqkv \\
        3 & 4 & 8 & 0.2 & Wqkv \\
        4 & 8 & 16 & 0.2 & Wqkv, Wi \\
        \bottomrule
    \end{tabular}
\end{table}

\begin{table}[h!]
    \caption{Training configuration per dataset. All use AdamW with weight decay 0.01, cosine annealing, 10\% warmup, gradient clipping at $L_2$ norm 1.0, and per-adapter bagging (80\% of problems).}
    \label{tab:training_config}
    \centering
    \resizebox{\textwidth}{!}{%
    \begin{tabular}{@{}lccccccc@{}}
        \toprule
        Dataset & Train problems & Candidates & Usable problems & Pairs & Batch size & Max length & Early stop patience \\
        \midrule
        GSM8K & 7{,}473 & 32 (multi) & 2{,}464 & 39{,}424 & 32 & 8{,}192 & 2 \\
        MuSR & 605 & 32 (multi) & 544 & 8{,}704 & 8 (MB) / 16 (DB) & 4{,}096 (MB) / 512 (DB) & 2 \\
        TravelPlanner & 175 & 32 (multi) & 175 & 2{,}800 & 32 & 8{,}192 & 2 \\
        TACO & 601 & 32 (multi) & 559 & 7{,}096 & 32 & 5{,}120 & 2 \\
        Knights \& Knaves & 6{,}200 & 32 (multi) & 4{,}426 & 70{,}816 & 16 & 1{,}536 & 3 \\
        \bottomrule
    \end{tabular}%
    }
\end{table}

\paragraph{Energy head architecture.}
Each adapter's energy head maps the \texttt{[CLS]} representation to a scalar energy:
$\text{LayerNorm}(d) \to \text{Linear}(d, d) \to \text{GELU} \to \text{Dropout}(0.2) \to \text{Linear}(d, 1)$,
where $d{=}768$ for both ModernBERT-base and DeBERTa-v3-base.
Trainable parameters per adapter: LoRA weights (${\sim}$0.3-1.2M depending on rank and target modules) plus energy head (${\sim}$0.6M), totalling ${\sim}$3\% of backbone parameters per member.
The full $K{=}5$ ensemble adds ${\sim}$15\% trainable overhead relative to the frozen backbone.

\paragraph{DeBERTa adapter configuration.}
DeBERTa-v3-base uses disentangled attention projections instead of fused QKV.
The heterogeneous adapters target:
adapter~0: \texttt{query\_proj, value\_proj};
adapter~1: \texttt{query\_proj, value\_proj, key\_proj};
adapter~2: \texttt{query\_proj, value\_proj} ($r{=}16$);
adapter~3: \texttt{query\_proj, value\_proj} ($r{=}4$);
adapter~4: \texttt{query\_proj, value\_proj, output\_proj}.

\paragraph{Training dynamics.}
On GSM8K (ModernBERT multi-model), all five adapters converge within 3-4 epochs, with best validation losses of 0.055-0.058.
On MuSR (ModernBERT multi-model), adapters train for 3-4 epochs with best validation losses of 0.64-0.65 (well below the random baseline of $\log 2 \approx 0.693$, confirming that a learnable signal is present despite the difficulty of the task).
DeBERTa adapters consistently train for more epochs (5 on MuSR, 2-5 on GSM8K) than ModernBERT, reflecting more stable optimisation at the shorter 512-token context.
All training uses fp16 automatic mixed precision on a single NVIDIA L40S (46\,GB); peak memory usage is ${\sim}$31\,GB at batch size 32.
Wall-clock time per adapter is ${\sim}$20 minutes per epoch on GSM8K and ${\sim}$5 minutes per epoch on MuSR.

\section{Training Data Details}
\label{app:training_data}
\vspace{-5pt}

\subsection{GSM8K}

Multi-model candidates: 7{,}473 problems $\times$ 32 candidates (8 per generator).
Classification by answer matching yields an overall candidate accuracy of 95\% (dominated by contamination).
Of 7{,}473 problems, 2{,}464 have both correct and incorrect candidates across the four generators (usable for contrastive training); 5{,}009 have all-correct pools (excluded).
LLaMA-3.1-8B contributes 1{,}309 unique negative problems (91.4\% accuracy), providing 4-10$\times$ more negatives than the other generators.
Contrastive pairs are capped at 16 per problem, yielding 39{,}424 total pairs split 80/20 at the problem level.

\subsection{MuSR}

756 problems split 80/20: 605 train, 151 test.
Multi-model candidates: 32 per problem (8 per generator, except Qwen which contributes 16 on some subsets).
Average candidate accuracy: 57\%.
Of 605 train problems, 544 (90\%) have both correct and incorrect candidates, providing 3$\times$ better coverage than GSM8K despite being 12$\times$ smaller.
This coverage advantage is the primary reason MuSR produces stronger training signal per problem.

\subsection{TravelPlanner}

175 train, 50 test problems from the validation split (sole-planning mode).
Multi-model candidates: 32 per problem.
Classification uses a continuous violation score in $[0, 1]$ aggregating 8 normalised constraint dimensions.
Positives are the top 25\% (8 of 32) per problem; negatives are the bottom 75\% (24 of 32).
All 175 train problems have both positives and negatives by construction.
Gemma-4-26B dominates with 67.9\% of its candidates in the positive set; LLaMA achieves only 8.6\%.

\subsection{TACO}

695 fully-clean problems from \texttt{likaixin/TACO-verified}, filtered to drop \texttt{VERY\_HARD} (0\% solve rate).
601 train, 94 test (88 test problems usable after filtering all-correct and all-incorrect pools).
Classification by automated test execution: candidates pass if every test case returns correct output within a 4-second timeout.
Positives capped at 8 per problem; negatives capped at 24, ranked by status severity.
Overall candidate accuracy: 31.9\% (train), with Gemma-4-26B at 84.6\% producing 58\% of all positives. A confounding analysis (\S\ref{app:confounding}) confirmed this imbalance caused a model-identity shortcut; DFR~\citep{kirichenko_2023_dfr} retraining on group-balanced data mitigates it.

\section{Dataset Details}
\label{app:datasets}
\vspace{-5pt}

\subsection{Exploratory Data Analysis}
\label{app:datasets:eda}

We characterise the four benchmarks along three axes: scale (split sizes and candidate counts),
length (problem and candidate token counts under the ModernBERT tokenizer),
and difficulty (a dataset-specific proxy: number of reasoning steps for GSM8K,
MuSR subset and number of choices, trip duration in days for TravelPlanner,
and official difficulty label for TACO).

\begin{table}[h!]
    \caption{Dataset overview. Candidate accuracy is the fraction of correct candidates across all generators.}
    \label{tab:dataset_overview}
    \centering
    \resizebox{\textwidth}{!}{%
    \begin{tabular}{@{}lrrrrccl@{}}
        \toprule
        Dataset & Train & Test & $N$ & Cand.\ accuracy & Label type & $\Econstraint$ & Primary metric \\
        \midrule
        GSM8K & 7{,}473 & 1{,}319 & 32 & 95\% & Binary (answer match) & None & pass@1 \\
        MuSR & 605 & 151 & 32 & 57\% & Binary (answer match) & None & pass@1 \\
        TravelPlanner & 175 & 50 & 32 & Top-25\% & Continuous (viol.\ score) & 5 constraints & Mean violation \\
        TACO & 601 & 88 & 32 & 32\% & Binary (execution) & None & pass@1 \\
        Knights \& Knaves & 6{,}200 & 700 & 32 & 58\% & Binary (exact match) & Statement consistency & pass@1 \\
        \bottomrule
    \end{tabular}%
    }
\end{table}

\subsection{Per-Model Candidate Quality}
\label{app:datasets:permodel}


\begin{table}[h!]
    \caption{Per-model candidate accuracy on the training set. Gemma-4-26B consistently dominates; LLaMA-3.1-8B contributes the most diverse errors.}
    \label{tab:permodel}
    \centering
    \begin{tabular}{@{}lccccc@{}}
        \toprule
        Model & GSM8K & MuSR & TravelPlanner & TACO & K\&K \\
        \midrule
        Gemma-4-26B-IT & 96.7\% & 66.7\% & 67.9\% & 84.6\% & 99.9\% \\
        GPT-OSS-20B & 96.1\% & 65.3\% & 9.8\% & 25.8\% & 99.0\% \\
        Qwen-2.5-7B & 95.1\% & 51.4\% & 13.6\% & 16.9\% & 19.7\% \\
        LLaMA-3.1-8B & 91.4\% & 49.7\% & 8.6\% & 10.5\% & 11.7\% \\
        \bottomrule
    \end{tabular}
\end{table}

On GSM8K, the high candidate accuracy (91-97\%) across all models reflects likely contamination.
On MuSR, the gap between Gemma (67\%) and Qwen/LLaMA (${\sim}$50\%) is substantial, providing the diversity of error patterns that drives contrastive learning.
On TACO, Gemma produces 58\% of all positive candidates. The initial scorer selected Gemma in 98\% of test problems; after DFR deconfounding, picks are distributed uniformly (22/24/34/20\%).

\subsection{GSM8K}
\label{app:gsm8k}

GSM8K~\citep{cobbe_2021_gsm8k} comprises 8{,}792 grade-school math word problems (7{,}473 train, 1{,}319 test).
Each problem is a short narrative ending with a question; the solution is a chain-of-thought derivation concluding with a numeric answer after the \texttt{\#\#\#\#} delimiter.
On GSM8K, $\Econstraint$ does not apply (no structured constraints beyond answer correctness), so the method reduces to distributional $\muq$ reranking only.

The primary metric is \emph{pass@1 after reranking}.
Secondary metrics include pass@$N$~\citep{chen_2021_codex} curves, energy separation (mean energy gap between correct and incorrect candidates), Kendall $\tau$ rank correlation, $\sigma$-AUROC, and expected calibration error (ECE;~\citealp{guo_2017_calibration}).

\paragraph{Contamination.}
GSM8K was released in 2021 and is widely included in LLM pretraining~\citep{yang_2023_rethink_benchmark}.
Random candidate accuracy of 92.7\% and self-consistency at 97.3\% are both implausibly high for grade-school math, suggesting that generators are reproducing memorised answers rather than reasoning.
We report GSM8K for completeness and backward comparability, but emphasise MuSR (released 2024, post-dating all generators' pretraining) as the clean signal.

\subsection{MuSR}
\label{app:musr}

MuSR~\citep{sprague_2024_musr} (Multi-step Soft Reasoning) consists of 756 narrative reasoning problems.
Three subsets test distinct reasoning capacities:
\emph{murder mysteries} (250 problems, 2 choices) require identifying a perpetrator from a narrative with multiple suspects;
\emph{object placements} (256 problems, 3 choices) require tracking object locations across a multi-paragraph narrative;
\emph{team allocation} (250 problems, 3 choices) require matching people to roles based on described traits.
Narratives span 3{,}000-6{,}000 characters and require multi-hop inference over entities and attributes.

$\Econstraint$ does not apply; the method reduces to distributional reranking.
The primary metric is pass@1.

\subsection{TravelPlanner}
\label{app:travelplanner}

TravelPlanner~\citep{xie_2024_travelplanner} is a multi-constraint planning benchmark.
Given a natural-language query specifying a trip (origin, destinations, dates, budget, preferences), the model produces a day-by-day JSON itinerary.
All experiments use \emph{sole-planning mode}: human-curated reference information is provided directly to the LLM.

We define five analytical constraint terms following the original benchmark categories~\citep{xie_2024_travelplanner}, each verified against the sandbox database:
\begin{itemize}[noitemsep]
    \item \emph{Budget}: $C_{\text{budget}} = \text{excess} / \text{budget}$, normalised to $[0, 1]$.
    \item \emph{Connectivity}: fraction of inter-city legs with invalid transport routes.
    \item \emph{Completeness}: fraction of missing required fields (accommodation, meals, attractions per day).
    \item \emph{Preferences}: fraction of unsatisfied user preferences.
    \item \emph{Diversity}: fraction of repeated entities across days.
\end{itemize}
Additionally, \emph{hallucination count} (entities not in the sandbox), \emph{structure violations} (route reasonableness), and \emph{parse success} are tracked.
The violation score is the mean of all 8 normalised dimensions.
The primary metric is mean violation score; secondary metric is pass@1 at $\theta{=}0$ (zero violations).

\paragraph{LLM-as-Judge candidate pool (GSM8K).}
LLM-as-Judge in \cref{tab:main_results} is evaluated on a slightly more lenient GSM8K candidate pool than the rest of the comparison: random pass@1 is 92.0\% (vs.\ 92.7\% on our pool) and oracle@32 is 99.2\% (vs.\ 98.5\%). The judge's headline 99.2\% essentially matches its pool's oracle, so the row reflects pool ceiling rather than a clean head-to-head with our scorer.

\begin{table}[!t]
    \caption{Qualitative examples from each benchmark. Each row shows a truncated problem, a candidate excerpt, and the verification outcome. Correct candidates receive low energy; incorrect candidates receive high energy.}
    \label{tab:examples}
    \centering
    \small
    \resizebox{\textwidth}{!}{%
    \begin{tabular}{@{}p{1.5cm}p{5.5cm}p{6.5cm}p{2.5cm}@{}}
        \toprule
        Dataset & Problem (truncated) & Candidate excerpt & Verdict \\
        \midrule
        GSM8K &
        Janet's ducks lay 16 eggs per day. She eats three for breakfast every morning and bakes muffins for her friends every day with four. She sells every remaining duck egg at the farmers' market for \$2 each. How much does she make every day? &
        Janet has 16 eggs per day. She eats 3 and bakes 4, so she uses $3 + 4 = 7$ eggs. She has $16 - 7 = 9$ eggs left. She sells them for \$2 each: $9 \times 2 = 18$. \newline \texttt{\#\#\#\# 18} &
        \textcolor{green!50!black}{Correct} \newline $E{=}{-}2.95$ \\
        \midrule
        MuSR &
        A murder mystery narrative (3{,}200 chars): three suspects at a dinner party. Motive, alibi, and physical evidence described across 6 paragraphs. ``Who is the most likely murderer?'' Choices: (A)~Alice, (B)~Bob, (C)~Carol. &
        Let me trace the evidence step by step. Alice has an alibi confirmed by two witnesses\ldots Bob was seen near the kitchen\ldots The knife found matches Bob's set\ldots \newline \texttt{\#\#\#\# 2} &
        \textcolor{green!50!black}{Correct (B)} \newline $E{=}{-}0.31$ \\
        \midrule
        TravelPlanner &
        Please provide a travel plan departing from Las Vegas to Denver for 3 days, March 12-14, 2022. Budget: \$1{,}600. Solo trip. &
        \texttt{[\{``days'': 1, ``current\_city'': ``from Las Vegas to Denver'', ``transportation'': ``Flight F3877057'', ``lunch'': ``The Urban Socialite, Denver'', ``dinner'': ``Kloof Street House, Denver'', ``accommodation'': ``Luxury Studio Suite\ldots''\}, \ldots]} &
        \textcolor{green!50!black}{0 violations} \newline viol.\ score: 0.00 \\
        \midrule
        TACO &
        A flower shop has $n$ bouquets with $a_i$ flowers each. Make the maximum number of large bouquets (two or more initial bouquets, odd total). Each initial bouquet used at most once. &
        \texttt{def solve(): n = int(input()); a = list(map(int, input().split())); odd = sum(1 for x in a if x \% 2); even = n - odd; print(min(odd, even) + max(0, (odd - even) // 2) if odd > even else odd)} &
        \textcolor{red!70!black}{Wrong answer} \newline test 1 failed \\
        \midrule
        K\&K &
        Oliver says ``Ethan is a knight.'' Ethan says ``Oliver is a knave or I am a knight.'' (2 characters) &
        If Oliver is a knight, his statement is true $\Rightarrow$ Ethan is a knight. Then Ethan's statement ``Oliver is a knave or I am a knight'' = ``F or T'' = T. Consistent. \newline \texttt{\{``Oliver'': ``knight'', ``Ethan'': ``knight''\}} &
        \textcolor{green!50!black}{Correct} \newline $E{=}{-}4.47$ \\
        \bottomrule
    \end{tabular}%
    }
\end{table}

\subsection{TACO}
\label{app:taco}

TACO~\citep{li_2023_taco} (Topics in Algorithmic Code generation) consists of competitive programming problems sourced from Codeforces, AtCoder, and related platforms.
We use the verified subset (\texttt{likaixin/TACO-verified}), filtered to drop \texttt{VERY\_HARD} problems (0\% solve rate).
Each problem includes a description, input/output format, constraints, and test cases.

Verification is execution-based: candidate code is extracted from the chain-of-thought response, executed against all test cases with a 4-second timeout per test, and labelled correct only if every test passes.
$\Econstraint$ does not apply; the binary execution label serves as both the training signal and the evaluation metric.
The primary metric is pass@1.

\paragraph{Contamination note.}
TACO problems are sourced from public competitive-programming sites.
Per-difficulty pass@1 (particularly \texttt{HARD}) is the cleanest signal for genuine reasoning; \texttt{EASY} problems are likely seen during pretraining.

\subsection{Knights and Knaves}
\label{app:kk}

Knights and Knaves (K\&K) puzzles~\citep{xie_2024_kk} are formal deductive reasoning problems.
Each puzzle presents $n$ characters (2-8), each either a knight (always tells the truth) or a knave (always lies), each making exactly one statement.
The task is to determine the unique assignment of knight/knave labels satisfying all statements simultaneously.

We use the dataset of \citet{xie_2024_kk} from HuggingFace (\texttt{K-and-K/knights-and-knaves}): 6{,}200 train and 700 test problems (100 per tier, 2ppl-8ppl).
Candidates are classified by exact match against the ground-truth boolean assignment.
A statement-consistency constraint checker independently verifies $\Econstraint{=}0 \Leftrightarrow \text{correct}$ on all 34{,}120 candidates checked, confirming perfect verification.

Token lengths are short (p50${\sim}$900 tokens for problem + candidate CoT); we use \texttt{max\_length=1536}.
Per-model accuracy reveals an extreme split: Gemma-4-26B and GPT-OSS-20B achieve ${\sim}$99\% accuracy, while Qwen-2.5-7B (19.7\%) and LLaMA-3.1-8B (11.7\%) rarely solve puzzles beyond 3 characters.
This split creates a strong model-identification confound in the multi-model setting (see \cref{sec:exp:results}).

\subsection{Qualitative Examples}
\label{app:examples}

\Cref{tab:examples} shows one problem-candidate pair per dataset, illustrating the input format, reasoning style, and verification signal.
The examples illustrate several properties of the benchmarks.
GSM8K candidates are short chain-of-thought derivations with a numeric answer delimiter; the verification signal is binary answer matching.
MuSR candidates involve multi-hop narrative reasoning over thousands of characters; correct answers require tracking entities and evidence across paragraphs.
TravelPlanner candidates are structured JSON itineraries verified against 8 constraint dimensions; even ``correct'' candidates (zero violations) may vary in quality, motivating the continuous scoring.
TACO candidates are executable Python code verified by automated test execution; the most common failure mode is \texttt{wrong\_answer} (56.6\% of negatives), where syntactically valid code produces incorrect output on edge cases.

\section{Extended Results and Analysis}
\label{app:extended}
\vspace{-5pt}

This section consolidates all extended results and analysis beyond the main paper.
\S\ref{app:ablations} isolates five design choices (ensemble size, pool size, multi-model training, backbone, energy decomposition).
\S\ref{app:energy} reports energy separation metrics and calibration diagnostics.
\S\ref{app:difficulty} breaks down per-difficulty performance on TACO, K\&K, and MuSR.
\S\ref{app:selective} evaluates selective prediction via $\sigma$-abstention.
\S\ref{app:twopass} analyses the two-pass $\sigma$-triage mechanism across datasets and pool configurations.
\S\ref{app:travelplanner_frontier} decomposes TravelPlanner violations per constraint dimension, comparing the Distributional EBM against frontier models.
\S\ref{app:transfer} reports the full zero-shot transfer matrix including DeBERTa backbones and MATH-500.
\S\ref{app:confounding} presents the six-phase TACO confounding analysis and DFR mitigation.

\subsection{Ablation Studies}
\label{app:ablations}

We isolate five design choices: ensemble size ($K$), candidate pool size ($N$), single-model vs.\ multi-model training, backbone architecture, and energy decomposition.
All ablations use ModernBERT unless noted; DeBERTa results are included where the 512-token context does not truncate candidates.

\begin{figure}[h]
    \centering
    \includegraphics[width=0.85\textwidth]{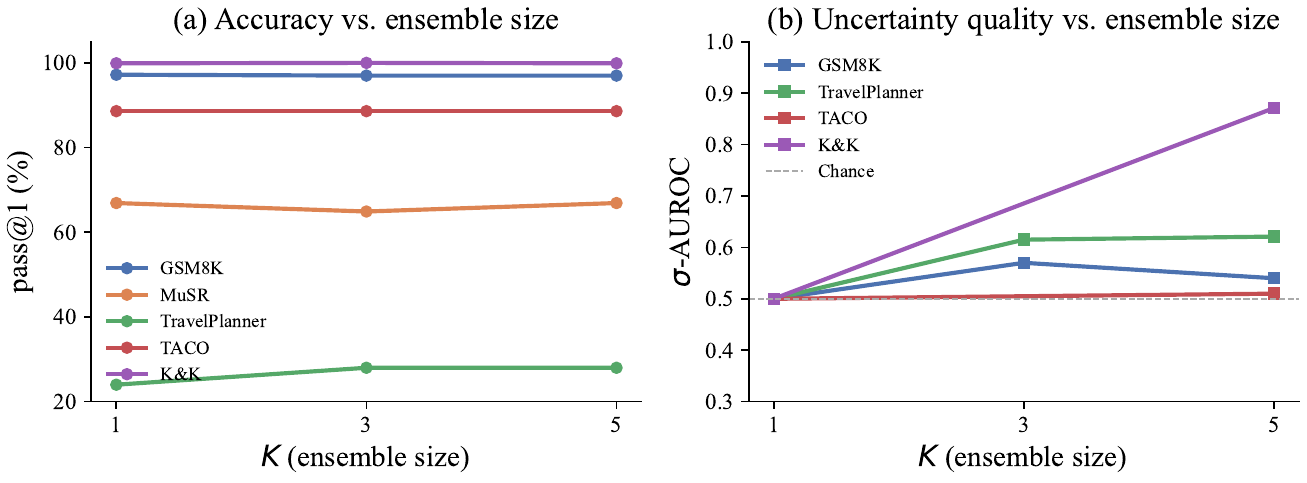}
    \caption{Effect of ensemble size $K$. \textbf{(a)}~Pass@1 is stable across $K$ on all datasets; the ensemble does not improve top-1 accuracy. \textbf{(b)}~$\sigma$-AUROC improves from chance (0.5 at $K{=}1$) to 0.87 on K\&K at $K{=}5$, confirming that structurally heterogeneous adapters produce informative uncertainty. The ensemble is justified by uncertainty estimation, not accuracy.}
    \label{fig:ensemble_k}
\end{figure}

\paragraph{Ensemble size (\cref{tab:full_k_sweep}, \cref{fig:ensemble_k}).}
Increasing $K$ from 1 to 5 has negligible effect on pass@1 across all datasets: GSM8K stays at 97.0-97.2\%, MuSR fluctuates between 64.9\% and 66.9\%, and TACO is constant at 88.6\%.
The one exception is TravelPlanner, where $K{\ge}3$ provides a +4\,pp lift (24.0\%$\to$28.0\%), likely because the continuous violation labels create noisier gradients that benefit from ensemble averaging.
The ensemble is justified not by accuracy but by uncertainty estimation: $\sigma$-AUROC is undefined at $K{=}1$ (single adapter has no variance) and improves to 0.87 at $K{=}5$ on K\&K.
DeBERTa shows the same pattern on GSM8K (96.4-96.6\% across $K$) and is not evaluated on TravelPlanner, TACO, or K\&K due to its 512-token context truncating candidates.

\begin{figure}[h]
    \centering
    \includegraphics[width=\textwidth]{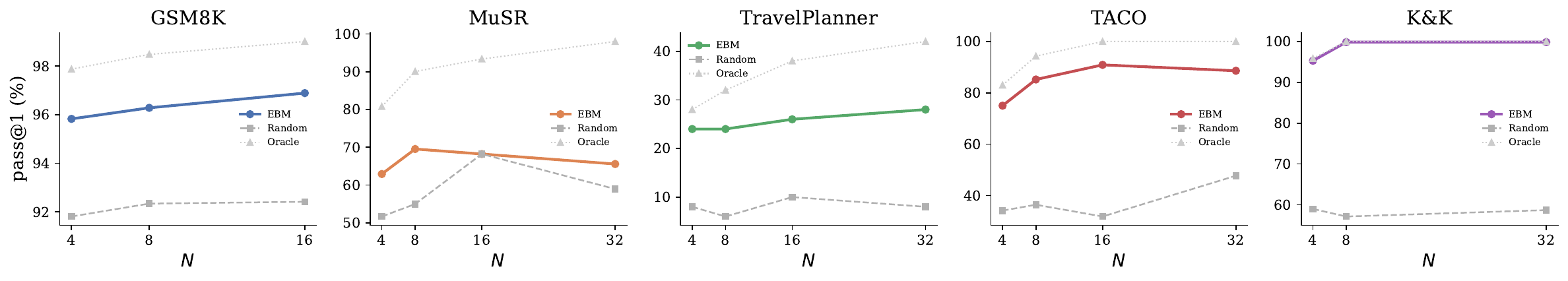}
    \caption{Candidate pool scaling. The Distributional EBM's advantage over random grows with $N$ on GSM8K; on MuSR the advantage peaks at $N{=}8$ and diminishes as the pool saturates.}
    \label{fig:pool_scaling}
\end{figure}

\paragraph{Candidate pool size (\cref{tab:full_pool}, \cref{fig:pool_scaling}).}
The Distributional EBM's advantage over random grows with pool size on binary-label tasks.
On GSM8K, the margin increases from +4.0\,pp at $N{=}4$ to +4.3\,pp at $N{=}32$.
On TACO, pass@1 scales from 75.0\% ($N{=}4$) to 88.6\% ($N{=}32$), closing 86\% of the random-to-oracle gap.
On MuSR, the advantage peaks at $N{=}8$ (+14.5\,pp over random) then diminishes at $N{=}16$ and $N{=}32$ as the pool becomes saturated with similar candidates, indicating diminishing returns on long-context reasoning tasks.

\paragraph{Single-model vs.\ multi-model (\cref{tab:single_multi}).}

\begin{table}[h!]
    \caption{Ensemble size ($K$) sweep across all datasets and both backbones. MB = ModernBERT, DB = DeBERTa. All multi model. TACO values are pre-DFR (the identity shortcut saturates at $K{=}1$, making the sweep uninformative for genuine quality discrimination; after DFR deconfounding, pass@1 is 88.6\%). DeBERTa (512-token context) is not evaluated on TravelPlanner, TACO, or K\&K because candidate outputs exceed its context window.} 
    \label{tab:full_k_sweep}
    \centering
    \begin{tabular}{@{}clccccc@{}}
        \toprule
        $K$ & Backbone & GSM8K & MuSR & TravelPlanner & TACO & K\&K \\
        \midrule
        1 & MB & 97.2\% & 66.9\% & 24.0\% & 92.0\% & 99.9\% \\
        3 & MB & 97.0\% & 64.9\% & 28.0\% & 92.0\% & 100.0\% \\
        5 & MB & 97.0\% & 66.9\% & 28.0\% & 92.0\% & 99.9\% \\
        \midrule
        1 & DB & 96.5\% & 62.3\% & -- & -- & -- \\
        3 & DB & 96.4\% & 63.6\% & -- & -- & -- \\
        5 & DB & 96.6\% & 62.9\% & -- & -- & -- \\
        \bottomrule
    \end{tabular}
\end{table}

\begin{table}[h!]
    \caption{Candidate pool size ($N$) ablation (ModernBERT, multi-model). \textbf{Bold} marks the best EBM result per dataset. TACO values are pre-DFR; after deconfounding the $N{=}32$ result is 88.6\%.}
    \label{tab:full_pool}
    \centering
    \begin{tabular}{@{}lcccccc@{}}
        \toprule
        & \multicolumn{2}{c}{GSM8K} & \multicolumn{2}{c}{MuSR} & \multicolumn{2}{c}{TACO} \\
        \cmidrule(lr){2-3}\cmidrule(lr){4-5}\cmidrule(lr){6-7}
        $N$ & EBM & Random & EBM & Random & EBM & Random \\
        \midrule
        4 & 95.8\% & 91.8\% & 62.9\% & 51.7\% & 75.0\% & 34.1\% \\
        8 & 96.3\% & 92.3\% & \textbf{69.5\%} & 55.0\% & 85.2\% & 36.4\% \\
        16 & 96.9\% & 92.4\% & 68.2\% & 68.2\% & 90.9\% & 31.8\% \\
        32 & \textbf{97.0\%} & 92.7\% & 65.6\% & 58.9\% & \textbf{92.0\%} & 47.7\% \\
        \bottomrule
    \end{tabular}
\end{table}

\begin{table}[h!]
    \caption{Single-model vs.\ multi-model comparison (ModernBERT, $K{=}5$). Single-model uses Qwen-2.5-7B only ($N{=}16$). Multi-model uses $N{=}16$ (4 models $\times$ 4 candidates); the main results (\cref{tab:main_results}) use $N{=}32$, explaining small differences (\textit{e.g.}, MuSR 66.9\% here vs.\ 67.7\%). \textbf{Bold} marks the better result per column.}
    \label{tab:single_multi}
    \centering
    \begin{tabular}{@{}lccccc@{}}
        \toprule
        & GSM8K & MuSR & TravelPlanner & TACO & K\&K \\
        \midrule
        Single-model EBM & 93.1\% & 53.6\% & 8.0\% & 36.4\% & 40.1\% \\
        Multi-model EBM & \textbf{97.0\%} & \textbf{66.9\%} & \textbf{28.0\%} & \textbf{88.6\%} & \textbf{99.9\%} \\
        \midrule
        $\Delta$ & +3.9\,pp & +13.3\,pp & +20.0\,pp & +52.2\,pp & +59.8\,pp \\
        \bottomrule
    \end{tabular}
\end{table}

Multi-model candidate diversity provides the largest gain of any design choice on every dataset (\cref{tab:single_multi}).
The effect is most dramatic on K\&K (+59.8\,pp) and TACO (+52.2\,pp), where the generator accuracy gap between the strongest and weakest model is largest.
On MuSR, single-model training (Qwen only, 53.6\%) barely exceeds random (51.0\%); multi-model reaches 66.9\% because the four generators provide both diverse error patterns for contrastive learning and higher oracle coverage (98.0\% vs.\ 86.8\%).
On GSM8K, the gain is smallest (+3.9\,pp) because all models are 91-97\% correct (contamination), leaving little room for multi-model diversity to help.

\subsection{Energy Separation and Calibration}
\label{app:energy}

\begin{figure}[!h]
    \centering
    \includegraphics[width=0.7\textwidth]{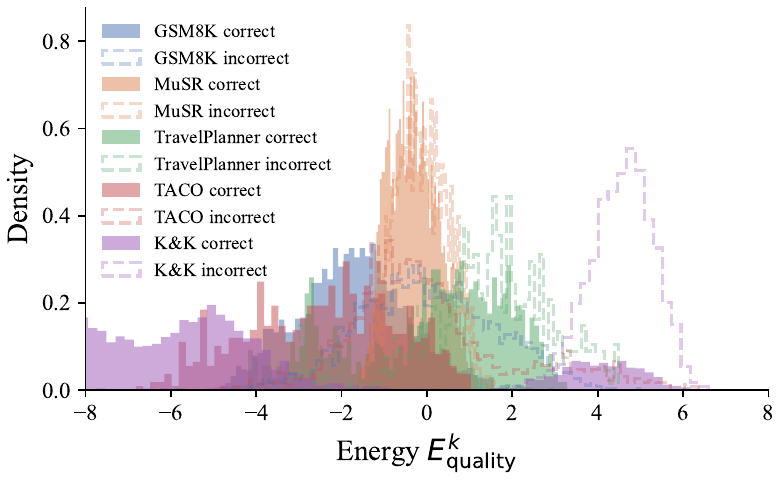}
    \caption{Energy distributions for correct (filled) and incorrect (dashed outline) candidates on GSM8K and MuSR (ModernBERT, multi-model). GSM8K shows clear bimodal separation; MuSR distributions overlap substantially, consistent with the smaller energy gap (0.19 vs.\ 1.76) and the difficulty of discriminating correct from incorrect narrative reasoning.}
    \label{fig:energy_ridgeplot}
\end{figure}

\begin{figure}[!h]
    \centering
    \includegraphics[width=\textwidth]{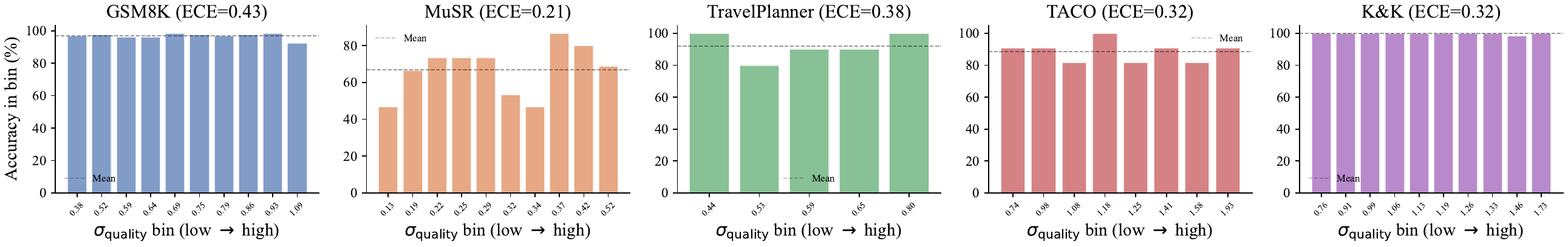}
    \caption{Calibration reliability diagrams. Bars show accuracy within each $\sigmaq$ bin. On GSM8K, accuracy is uniformly high across all bins (ECE = 0.43, the scorer is poorly calibrated). On MuSR, accuracy fluctuates non-monotonically with $\sigmaq$, confirming that the uncertainty signal is not reliably informative on this task.}
    \label{fig:calibration}
\end{figure}

\Cref{tab:energy_sep} and \cref{fig:calibration} report diagnostic metrics for the Distributional EBM across all five benchmarks; \cref{fig:energy_ridgeplot} visualises the correct and incorrect energy distributions.

K\&K exhibits the largest energy gap (9.00) and highest Kendall $\tau$ (0.62), reflecting the clean binary signal from the statement-consistency verifier and the high candidate diversity.
TACO's gap (2.49) and $\tau$ (0.44) are reported after DFR deconfounding; the pre-DFR values (4.30, 0.78) were inflated by the model-identity shortcut.
MuSR has the smallest gap (0.19) and lowest $\tau$ (0.14), consistent with the difficulty of discriminating correct from incorrect narrative reasoning where both candidates contain plausible multi-hop inferences.
TravelPlanner achieves the highest $\tau$ (0.49) despite a moderate gap (2.00), because its continuous violation labels produce a smoother energy landscape than binary correctness labels.
$\sigma$-AUROC ranges from 0.44 (MuSR, anti-correlated) to 0.87 (K\&K), confirming that ensemble uncertainty is task-dependent: reliable on tasks with clear correct/incorrect boundaries, unreliable on nuanced reasoning.
ECE is moderate across all datasets (0.21-0.43), with MuSR the best-calibrated (0.21) and GSM8K the worst (0.43).

\begin{table}[h!]
    \caption{Energy separation metrics (ModernBERT, multi-model, $K{=}5$). Energy gap = $\bar{E}(\text{incorrect}) - \bar{E}(\text{correct})$; positive means the scorer correctly assigns lower energy to correct candidates. \textbf{Bold} marks the best value per metric (higher is better for gap, $\tau$, AUROC; lower for ECE).}
    \label{tab:energy_sep}
    \centering
    \begin{tabular}{@{}lccccc@{}}
        \toprule
        Metric & GSM8K & MuSR & TravelPlanner & TACO & K\&K \\
        \midrule
        $\bar{E}(\text{correct})$ & $-$1.71 & $-$0.32 & $-$1.26 & $-$2.44 & $-$4.47 \\
        $\bar{E}(\text{incorrect})$ & 0.04 & $-$0.12 & 0.74 & 0.04 & 4.52 \\
        Energy gap & 1.76 & 0.19 & 2.00 & 2.49 & \textbf{9.00} \\
        Kendall $\tau$ & 0.27 & 0.14 & 0.49 & 0.44 & \textbf{0.62} \\
        $\sigma$-AUROC & 0.54 & 0.44 & 0.62 & 0.51 & \textbf{0.87} \\
        ECE & 0.43 & \textbf{0.21} & 0.28 & 0.32 & 0.32 \\
        \bottomrule
    \end{tabular}
\end{table}

\subsection{Per-Difficulty Analysis}
\label{app:difficulty}

This section breaks down results by task difficulty to reveal where the Distributional EBM excels and where it struggles.

\paragraph{TACO (\cref{tab:taco_difficulty}).}
The deconfounded EBM degrades gracefully with difficulty: 93.8\% on EASY, 90.9\% on MEDIUM, and 80.0\% on HARD.
It matches Sonnet~4.6 on MEDIUM and HARD problems, and substantially outperforms Qwen-72B at every difficulty level (by 16-60\,pp).
Opus~4.6 achieves 100\% across all tiers, reflecting the advantage of a much larger model on competitive programming.

\begin{table}[h!]
    \caption{TACO pass@1 by difficulty (multi-model). The EBM matches or exceeds Sonnet~4.6 on MEDIUM problems. $n$ = number of test problems.}
    \label{tab:taco_difficulty}
    \centering
    \begin{tabular}{@{}lccccc@{}}
        \toprule
        Difficulty & EBM & Opus 4.6 & Sonnet 4.6 & Qwen-72B & $n$ \\
        \midrule
        EASY & 93.8\% & 100.0\% & 100.0\% & 77.1\% & 48 \\
        MEDIUM & 90.9\%  & 100.0\% & 90.9\% & 45.5\% & 11 \\
        MEDIUM\_HARD & 81.8\% & 100.0\% & 90.9\% & 27.3\% & 11 \\
        HARD & 80.0\% & 100.0\% & 80.0\% & 20.0\% & 5 \\
        \bottomrule
    \end{tabular}
\end{table}

\paragraph{Knights and Knaves (\cref{tab:kk_difficulty}).}
The multi-model EBM achieves 100\% pass@1 on all tiers except 8-character puzzles (99.0\%), demonstrating that the scorer maintains near-perfect discrimination even as combinatorial complexity grows exponentially.
Gains over greedy scale with difficulty: +32\,pp on 2ppl to +55\,pp on 7ppl.
The single-model EBM (Qwen-only) shows a different pattern: gains are largest on the hardest tiers (+23.1\,pp on 8ppl, +14.5\,pp on 6ppl), suggesting that the scorer adds the most value precisely where the generator is weakest.
Qwen-72B degrades steeply from 75\% (2ppl) to 10\% (8ppl), falling below random (58.7\%) from 5ppl onward.

\begin{table}[h!]
    \caption{K\&K pass@1 by puzzle size (number of characters). Multi-model EBM achieves 100\% on all tiers except 8ppl. \textbf{Bold} marks the best result per tier. Single-model EBM gains are largest on the hardest tiers.}
    \label{tab:kk_difficulty}
    \centering
    \resizebox{0.85\textwidth}{!}{%
    \begin{tabular}{@{}lcccccccc@{}}
        \toprule
        & \multicolumn{3}{c}{Multi-model ($N{=}32$)} & \multicolumn{3}{c}{Single-model ($N{=}16$)} & \multicolumn{2}{c}{Frontier} \\
        \cmidrule(lr){2-4}\cmidrule(lr){5-7}\cmidrule(lr){8-9}
        Tier & Greedy & EBM & Lift & Greedy & EBM & Lift & Qwen-72B & Random \\
        \midrule
        2ppl & 68.0\% & \textbf{100.0\%} & +32.0\,pp & 58.7\% & 58.7\% & +0.0\,pp & 75.0\% & 58.7\% \\
        3ppl & 56.0\% & \textbf{100.0\%} & +44.0\,pp & 45.7\% & 48.9\% & +3.2\,pp & 63.0\% & 58.7\% \\
        4ppl & 60.0\% & \textbf{100.0\%} & +40.0\,pp & 25.8\% & 38.7\% & +12.9\,pp & 48.0\% & 58.7\% \\
        5ppl & 59.0\% & \textbf{100.0\%} & +41.0\,pp & 28.2\% & 29.5\% & +1.3\,pp & 32.0\% & 58.7\% \\
        6ppl & 55.0\% & \textbf{100.0\%} & +45.0\,pp & 17.7\% & 32.3\% & +14.5\,pp & 29.0\% & 58.7\% \\
        7ppl & 45.0\% & \textbf{100.0\%} & +55.0\,pp & 14.6\% & 19.5\% & +4.9\,pp & 16.0\% & 58.7\% \\
        8ppl & 52.0\% & \textbf{99.0\%} & +47.0\,pp & 10.3\% & 33.3\% & +23.1\,pp & 10.0\% & 58.7\% \\
        \bottomrule
    \end{tabular}%
    }
\end{table}

\paragraph{MuSR (\cref{tab:musr_subsets}).}
MuSR's three subsets differ in structure: murder mysteries are 2-choice (50\% random baseline), while object placements and team allocation are 3-choice (33\% random).
Average candidate accuracy is highest on murder mysteries (62.2\%) and lowest on object placements (52.1\%), reflecting the difficulty of spatial reasoning relative to narrative deduction.
The EBM's gains are largest on the 3-choice subsets where the random baseline is lower and there is more room for discrimination.

\begin{table}[h!]
    \caption{MuSR per-subset accuracy (multi-model). Murder mysteries are 2-choice (50\% random); objects and team are 3-choice (33\% random).}
    \label{tab:musr_subsets}
    \centering
    \begin{tabular}{@{}lccc@{}}
        \toprule
        Subset & Avg.\ candidate accuracy & Random baseline & $n$ choices \\
        \midrule
        Murder mysteries & 62.2\% & 50\% & 2 \\
        Object placements & 52.1\% & 33\% & 3 \\
        Team allocation & 57.2\% & 33\% & 3 \\
        \bottomrule
    \end{tabular}
\end{table}

\subsection{Selective Prediction via $\sigma$-Abstention}
\label{app:selective}

\Cref{tab:selective} reports pass@1 on the retained set after abstaining on the $X$\% of problems with highest $\sigmaq$.
On K\&K, selective prediction reaches 100\% by 20\% abstention, reflecting the strong $\sigma$-AUROC (0.87): the ensemble reliably identifies the single wrong prediction.
On TravelPlanner, abstaining on 40\% of problems raises pass@1 from 28.0\% to 36.7\% (+8.7\,pp), confirming that $\sigmaq$ captures meaningful quality uncertainty on continuous-label tasks.
On GSM8K, selective prediction is flat (97.0-97.4\%) because the model is already near-ceiling: there are only ${\sim}$40 wrong predictions out of 1{,}319.
On MuSR, accuracy \emph{decreases} with abstention (66.9\%$\to$61.9\% at 30\%), consistent with the anti-correlated $\sigma$-AUROC (0.44): the ensemble is more confident on wrong predictions than right ones, so abstaining on high-$\sigmaq$ problems removes easy ones.
On TACO (after DFR), selective prediction provides only +2.3\,pp at 50\% abstention (88.6\%$\to$90.9\%), down from +5.3\,pp pre-DFR, because the pre-DFR $\sigma$ was capturing model-identity uncertainty rather than quality uncertainty.

\begin{table}[h!]
    \caption{Selective prediction: pass@1 on retained problems after abstaining on the most uncertain (highest $\sigmaq$) fraction. \textbf{Bold} marks where a dataset reaches 100\% accuracy. TACO values are post-DFR; selective prediction is weak (+2.3\,pp at 50\% abstention) because the deconfounded $\sigma$ no longer captures model-identity uncertainty. MuSR pass@1 fluctuates non-monotonically with abstention (61.9--66.9\%); the anti-correlated $\sigma$-AUROC (0.44) means high-$\sigma$ problems are not preferentially incorrect, so abstention does not concentrate on the wrong predictions.}
    \label{tab:selective}
    \centering
    \begin{tabular}{@{}lccccc@{}}
        \toprule
        Abstain \% & GSM8K (MB) & MuSR (MB) & TravelPlanner & TACO & K\&K \\
        \midrule
        0\% & 97.0\% & 66.9\% & 28.0\% & 88.6\% & 99.9\% \\
        10\% & 97.4\% & 66.7\% & 28.9\% & 88.6\% & 99.8\% \\
        20\% & 97.3\% & 65.0\% & 30.0\% & 88.6\% & \textbf{100.0\%} \\
        30\% & 97.2\% & 61.9\% & 31.4\% & 88.5\% & \textbf{100.0\%} \\
        40\% & 97.2\% & 64.4\% & 36.7\% & 90.4\% & \textbf{100.0\%} \\
        50\% & 97.1\% & 66.7\% & 36.0\% & 90.9\% & \textbf{100.0\%} \\
        \bottomrule
    \end{tabular}
\end{table}

\subsection{Two-Pass $\sigma$-Triage Analysis}
\label{app:twopass}

\Cref{alg:inference} provides the full inference pseudocode for the two-pass $\sigma$-triage mechanism described in \cref{sec:method:twopass}.

\begin{algorithm}[h!]
\caption{Inference with $\sigma$-triage}
\label{alg:inference}
\begin{algorithmic}[1]
\REQUIRE Problem $x$, frozen LLMs $\{G_1, \ldots, G_M\}$, ensemble $\{f_k\}_{k=1}^K$, constraint set $\{C_j\}$, thresholds $\theta_\sigma$, $\theta_{\text{abstain}}$, weight $\lambda$
\ENSURE Selected candidate $y^*$ or \textsc{Abstain}
\STATE \textbf{Pass 1: Generate.} For each LLM $G_m$, sample $N_m$ candidates via temperature sampling. Pool: $\mathcal{Y}_1 = \{y_1, \ldots, y_{N_1}\}$.
\FOR{each $y_i \in \mathcal{Y}_1$}
    \STATE Parse $y_i$. If parse fails, set $\Econstraint(x, y_i) = +\infty$.
    \STATE Score with all $K$ adapters: $\{E^1(x, y_i), \ldots, E^K(x, y_i)\}$.
    \STATE $\muq(x, y_i) \gets \text{mean}(\{E^k\})$, \quad $\sigmaq(x, y_i) \gets \text{std}(\{E^k\})$.
    \STATE $\Econstraint(x, y_i) \gets \sum_j w_j \cdot C_j(x, y_i)$.
    \STATE $E(x, y_i) \gets \muq(x, y_i) + \lambda \cdot \Econstraint(x, y_i)$.
\ENDFOR
\STATE $y^* \gets \arg\min_{y_i} E(x, y_i)$.
\IF{$\Econstraint(x, y^*) > 0$ \textbf{or} $\overline{\sigmaq} > \theta_\sigma$}
    \STATE \textbf{Pass 2: Regenerate.} Format constraint violations of $y^*$ as feedback.
    \STATE Generate $N_2$ candidates conditioned on feedback: $\mathcal{Y}_2$.
    \STATE Score all $y_i \in \mathcal{Y}_2$ (repeat lines 3-7). Pool: $\mathcal{Y} = \mathcal{Y}_1 \cup \mathcal{Y}_2$.
    \STATE $y^* \gets \arg\min_{y_i \in \mathcal{Y}} E(x, y_i)$.
\ENDIF
\IF{$\sigmaq(x, y^*) > \theta_{\text{abstain}}$}
    \STATE \textbf{return} \textsc{Abstain}
\ELSE
    \STATE \textbf{return} $y^*$
\ENDIF
\end{algorithmic}
\end{algorithm}

\Cref{tab:twopass} reports two-pass $\sigma$-triage results across all datasets and pool configurations.
The mechanism helps when three conditions hold: (a)~pass-1 has headroom (not already near-ceiling), (b)~the feedback is structured enough to guide regeneration, and (c)~the generator can act on it.
TravelPlanner single-model satisfies all three: constraint violations (budget overruns, missing meals, invalid routes) provide concrete feedback, yielding +6.0\,pp.
TACO multi-model (confounded) fails catastrophically ($-$89.8\,pp) because all 88 problems are triaged and pass-2 candidates disrupt the Gemma-centric energy landscape.
After DFR deconfounding, TACO two-pass is neutral (88.6\%$\rightarrow$88.6\%) because the deconfounded scorer evaluates each candidate on quality alone, so re-pooling low-quality pass-2 candidates does not corrupt the ranking.
GSM8K two-pass reduces pass@1 by 1.7\,pp because doubt-inducing feedback causes the contaminated generator to second-guess its (memorised) correct answers.

\begin{table}[h!]
    \caption{Two-pass $\sigma$-triage results across datasets. $\theta_\sigma{=}0.8$ unless noted, $N_2{=}8$ pass-2 candidates from Qwen-2.5-7B with constraint/error feedback. \textbf{Bold} marks the only configuration where two-pass improves pass@1. ``--'' = not applicable (no problems triaged or no pass-2 candidates adopted).}
    \label{tab:twopass}
    \centering
    \resizebox{\textwidth}{!}{%
    \begin{tabular}{@{}llccccc@{}}
        \toprule
        Dataset & Pool & Triage rate & Pass-1 & Two-pass & $\Delta$ & Pass-2 adopted \\
        \midrule
        GSM8K & Single (MB v1) & 201/1319 (15\%) & 93.1\% & 91.4\% & $-$1.7\,pp & 56/201 \\
        MuSR & Multi & 0/151 (0\%) & 67.7\% & 67.7\% & 0.0\,pp & -- \\
        TravelPlanner & Single & 18/50 (36\%) & 74.0\% & \textbf{80.0\%} & +6.0\,pp & 9/18 \\
        TravelPlanner & Multi & 17/50 (34\%) & 92.0\% & 92.0\% & 0.0\,pp & 1/17 \\
        TACO (pre-DFR) & Multi & 88/88 (100\%) & 92.0\% & 2.3\% & $-$89.8\,pp & 0/88 \\
        TACO (DFR) & Multi & 84/88 (95.5\%) & 88.6\% & 88.6\% & 0.0\,pp & -- \\
        TACO & Single & 0/88 (0\%) & 36.4\% & 36.4\% & 0.0\,pp & -- \\
        K\&K & Single ($\theta_\sigma{=}0.5$) & 396/499 (79.4\%) & 85.2\% & 85.0\% & $-$0.2\,pp & 29/396 \\
        K\&K & Multi ($\theta_\sigma{=}0.5$) & 700/700 (100\%) & 99.9\% & 99.9\% & 0.0\,pp & 0/700 \\
        \bottomrule
    \end{tabular}%
    }
\end{table}

Two-pass helps when pass-1 has headroom, feedback is structured (\textit{e.g.}, specific constraint violations), and the generator can act on it.
TravelPlanner single-model satisfies all three conditions.
On K\&K single-model ($\theta_\sigma{=}0.5$), two-pass is near-neutral ($-$0.2\,pp overall) but shows a tier-dependent pattern: it helps on easy puzzles (2-4ppl: +2 to +4\,pp) where constraint feedback gives actionable corrections, but hurts on hard puzzles (7-8ppl: $-$10 to $-$13\,pp) where pass-2 candidates look confident to the ensemble but are more often wrong.
On K\&K multi-model, all 700 problems trigger triage (mean $\sigmaq{=}0.931$) due to high inter-model disagreement, but only 1 problem has a constraint violation receiving actionable pass-2 feedback; the remaining 699 already satisfy all constraints, so two-pass has zero effect (99.9\%$\to$99.9\%).
TACO multi-model fails catastrophically because: (a) all 88 problems trigger triage ($\sigma_{\text{pool}} > 0.8$ due to high inter-model disagreement), (b) pass-2 candidates from Qwen-7B with opaque error feedback are uniformly poor, and (c) re-pooling these low-quality candidates disrupts the energy landscape established by the strong pass-1 pool.

\paragraph{When does $\sigma$-triage help?}
Two patterns are visible. \emph{(a) Constraint-checkability.} $\sigma$-triage adds value where regeneration receives \emph{actionable} feedback: TravelPlanner's per-constraint violation messages (budget over by \$X, missing meal on day 2) translate directly into pass-2 prompts (+6.0\,pp); TACO's opaque \texttt{wrong\_answer} and doubt-inducing feedback on contaminated GSM8K do not. \emph{(b) Quality-distribution multimodality.} Triage benefits tasks where candidates cluster into discrete modes the ensemble can disagree about (K\&K's statement-consistency states); on tasks where candidates lie on a continuous quality manifold (MuSR's narrative inferences), $\sigma$ measures presentation variation rather than correctness. $\sigma$-triage therefore selects which regime each task falls into rather than applying a one-size-fits-all loop.

\subsection{TravelPlanner: Per-Constraint Frontier Comparison}
\label{app:travelplanner_frontier}

\Cref{tab:tp_perdim} decomposes the TravelPlanner violation score by constraint dimension, revealing where the Distributional EBM and frontier models have complementary strengths.
Frontier models (Opus, Sonnet, Qwen-72B) systematically fail on connectivity (24-28\% satisfaction) because they hallucinate transport routes not in the reference database.
The analytical constraint checker verifies routes against the database directly, giving the EBM+$\Econstraint$ system near-perfect connectivity (86\% satisfaction).
Conversely, frontier models outperform on budget (Opus: 0.040 vs.\ EBM+$\Econstraint$: 0.030), reflecting stronger numerical reasoning.
Preferences are the hardest dimension for all methods, including oracle (0.180 mean violation).

\begin{table}[h!]
    \caption{Per-violation-dimension mean scores on TravelPlanner (50 test problems). Lower is better. \textbf{Bold} marks the best (lowest) value per row.}
    \label{tab:tp_perdim}
    \centering
    \resizebox{\textwidth}{!}{%
    \begin{tabular}{@{}lcccccccc@{}}
        \toprule
        Dimension & Oracle & $\Econstraint$ only & EBM+$\Econstraint$ & EBM & Opus 4.6 & Sonnet 4.6 & Qwen-72B & Random \\
        \midrule
        Budget & 0.018 & \textbf{0.012} & 0.030 & 0.176 & 0.040 & 0.078 & 0.062 & 0.238 \\
        Completeness & 0.009 & 0.011 & 0.031 & \textbf{0.003} & 0.086 & 0.048 & 0.110 & 0.260 \\
        Hallucination & 0.005 & 0.011 & 0.011 & \textbf{0.002} & 0.015 & 0.037 & 0.070 & 0.119 \\
        Connectivity & \textbf{0.000} & 0.003 & 0.058 & 0.062 & 0.418 & 0.393 & 0.375 & 0.232 \\
        Preferences & \textbf{0.180} & 0.187 & 0.213 & 0.220 & 0.213 & 0.213 & 0.280 & 0.387 \\
        Diversity & 0.011 & 0.080 & 0.022 & \textbf{0.000} & 0.007 & 0.054 & 0.074 & 0.148 \\
        \bottomrule
    \end{tabular}%
    }
\end{table}

Frontier models systematically fail on connectivity (24-28\% satisfaction) because they hallucinate transport routes not present in the reference database.
The analytical constraint checker verifies routes against the database directly, giving the EBM+$\Econstraint$ system a structural advantage on this dimension.
Conversely, frontier models outperform on budget (Opus 4.6: 0.040 vs.\ EBM+$\Econstraint$: 0.030), reflecting their stronger numerical reasoning.
Preferences are the hardest dimension for all methods, including oracle (0.180 mean violation even for the best candidate).

\paragraph{Independent LLM-judge validation.}
To check that the violation score reflects genuine plan quality rather than artefacts of the constraint checker, two independent LLM judges (Opus~4.6 and Sonnet~4.6~\citep{anthropic_2026_claude}, $T{=}0$) score 120 candidates on a 1--5 scale using the original TravelPlanner criteria~\citep{xie_2024_travelplanner}. The judges agree strongly with each other (Spearman $\rho{=}0.67$, $p < 10^{-16}$) and both correlate negatively with the violation score (Opus: $\rho{=}{-}0.46$; Sonnet: $\rho{=}{-}0.38$; both $p < 10^{-5}$). Opus~4.6 rates EBM-selected plans higher than oracle plans on average (3.86 vs.\ 3.67), suggesting the learned scorer captures holistic coherence beyond what constraints alone measure.

\subsection{Zero-Shot Transfer: Full Matrix}
\label{app:transfer}

\begin{figure}[h]
    \centering
    \includegraphics[width=0.7\textwidth]{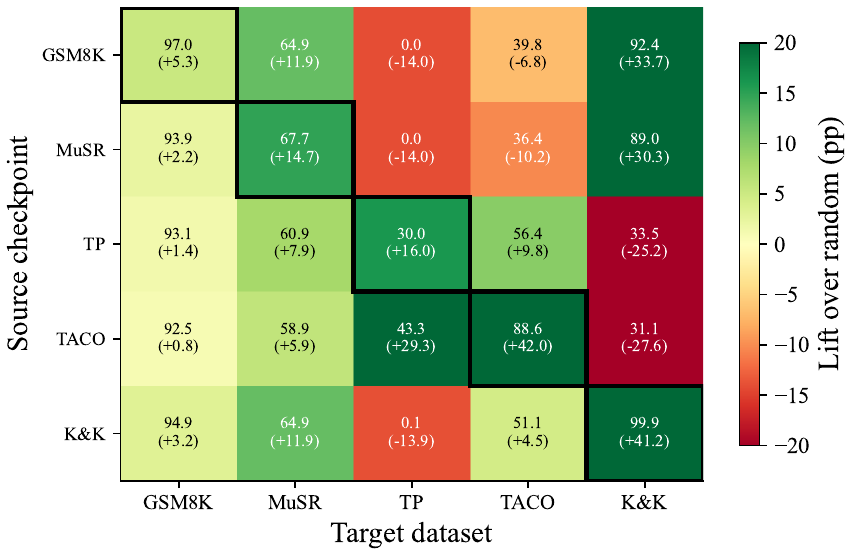}
    \caption{Zero-shot transfer heatmap (ModernBERT): each cell shows pass@1 and (lift over random). Colour encodes lift: green = positive transfer, red = negative. Black borders mark in-domain entries.}
    \label{fig:transfer_heatmap}
\end{figure}

\Cref{tab:full_transfer} reports the full zero-shot transfer matrix, including both ModernBERT and DeBERTa checkpoints and the MATH-500 benchmark.
\Cref{fig:transfer_heatmap} visualises the transfer asymmetry as lift over random.

Four patterns are evident.
First, \emph{hard-to-easy reasoning transfer works}: both GSM8K and MuSR checkpoints achieve 89-92\% on K\&K, and the MuSR-trained DeBERTa achieves 96.1\% on GSM8K, matching in-domain self-consistency.
Second, \emph{easy-to-hard reasoning transfer is weak}: the GSM8K checkpoint reaches only 64.9\% on MuSR (vs.\ 67.7\% in-domain), because the contaminated training data provides too few informative negatives.
Third, \emph{reasoning-to-code transfer is negative}: both GSM8K (39.8\%) and MuSR (36.4\%) checkpoints score \emph{below} random (46.6\%) on TACO, indicating that reasoning-quality signals actively mislead on code.
Fourth, \emph{no checkpoint transfers to MATH-500}: all achieve $\le$68\% vs.\ self-consistency at 72.6\%, likely due to the format mismatch (\texttt{\textbackslash boxed\{\}} vs.\ \texttt{\#\#\#\#}).

\begin{table}[!t]
    \caption{Full zero-shot transfer matrix (EBM pass@1, ModernBERT multi-model unless noted). \colorbox{gray!15}{Shaded cells} mark in-domain (diagonal) entries. \textbf{Bold} highlights notable transfer results. Baselines for reference: self-consistency (SC) and random. }
    \label{tab:full_transfer}
    \centering
    \small
    \resizebox{0.8\textwidth}{!}{%
    \begin{tabular}{@{}lcccccc@{}}
        \toprule
        Source $\rightarrow$ Target & GSM8K & MuSR & MATH-500 & TP & TACO & K\&K \\
        \midrule
        GSM8K (MB) & \cellcolor{gray!15}97.0\% & 64.9\% & 65.6\% & 0.0\% & 39.8\% & 92.4\% \\
        GSM8K (DB) & \cellcolor{gray!15}97.1\% & 63.6\% & 62.7\% & 0.0\% & 45.5\% & 68.4\% \\
        MuSR (MB) & 93.9\% & \cellcolor{gray!15}67.7\% & 67.6\% & 0.0\% & 36.4\% & 89.0\% \\
        MuSR (DB) & \textbf{96.1\%} & \cellcolor{gray!15}62.9\% & 67.8\% & 0.0\% & 39.8\% & 88.4\% \\
        TP (MB) & 93.1\% & 60.9\% & 67.2\% & \cellcolor{gray!15}30.0\% & 56.4\% & 33.5\% \\
        TACO (MB) & 92.5\% & 58.9\% & 67.4\% & 43.3\% & \cellcolor{gray!15}88.6\% & 31.1\% \\
        K\&K (MB) & 94.9\% & 64.9\% & 0.0\% & 0.136$^\diamond$ & 51.1\% & \cellcolor{gray!15}\textbf{99.9\%} \\
        \midrule
        Self-consistency & 96.1\% & 64.4\% & 72.6\% & 12.0\% & 36.4\% & 56.4\% \\
        Random & 91.7\% & 53.0\% & 68.4\% & 14.0\% & 46.6\% & 58.7\% \\
        \bottomrule
    \end{tabular}%
    }
\end{table}

Key observations:
(i) MuSR$\rightarrow$GSM8K transfer (96.1\% with DeBERTa) matches in-domain self-consistency, confirming that hard-data training produces transferable quality signals.
(ii) No checkpoint transfers positively to MATH-500 (all $\le$ 67.8\%, below SC at 72.6\%), likely due to format mismatch (\texttt{\textbackslash boxed\{\}} vs.\ \texttt{\#\#\#\#}).
(iii) Reasoning$\rightarrow$code transfer is negative: GSM8K (39.8\%) and MuSR (36.4\%) both score below TACO random (46.6\%), indicating that reasoning-quality signals actively mislead on code.
(iv) TACO$\rightarrow$TravelPlanner shows strong positive transfer (43.3\% vs.\ 14.0\% random, +29.3\,pp), and TravelPlanner$\rightarrow$TACO also transfers positively (56.4\% vs.\ 46.6\% random, +9.8\,pp), suggesting that execution-based labels produce more generalisable quality signals than reasoning-based labels.
(v) DeBERTa backbone transfer to K\&K is asymmetric: MuSR(DB)$\rightarrow$K\&K retains 88.4\% ($-$0.6\,pp vs.\ ModernBERT), while GSM8K(DB)$\rightarrow$K\&K drops to 68.4\% ($-$24.0\,pp). The MuSR-trained scorer learns quality signals robust to 512-token truncation, whereas the GSM8K scorer, trained on contaminated data with 95\% candidate accuracy, relies on shallow features that are lost when inputs are truncated.

\paragraph{Cross-pool transfer (Qwen-train $\rightarrow$ multi-test).}
\label{app:transfer:cross_pool}
A separate transfer axis tests whether the verifier generalises across \emph{generator distributions}: we take the Qwen-only-trained scorer (the same checkpoint used in the single-model row of \cref{tab:single_multi}) and evaluate it on the multi-model candidate test pool used for the headline results. \Cref{tab:cross_pool} reports the result. On MuSR, the cross-pool scorer reaches 69.5\% pass@1 (151 problems), \emph{matching} the multi-trained scorer (67.7\%) and exceeding the single-model in-domain configuration (53.6\%) by +15.9\,pp; the Qwen-only training distribution still produces a ranker that handles candidates from LLaMA, Gemma, and GPT-OSS without seeing them during training. On GSM8K, the cross-pool scorer reaches 90.5\% pass@1 (1{,}319 problems), below the multi-trained 97.0\% and even below pool-random (92.7\%); GSM8K's near-saturated candidate accuracy (95\%) leaves few informative negatives during Qwen-only training, so the scorer learns shallow signals that misrank the multi-model pool. This mirrors pattern (iv): contaminated training data produces non-transferable signals.

\begin{table}[h!]
    \caption{Cross-pool transfer: Qwen-only-trained EBM evaluated on the multi-model candidate test pool. \emph{Single-train, single-test} repeats the entries from the corresponding row of \cref{tab:single_multi} for context. All numbers are pass@1 with $N{=}32$ candidates per problem; multi-test pool is the same as the headline evaluation.}
    \label{tab:cross_pool}
    \centering
    \small
    \begin{tabular}{@{}lcc@{}}
        \toprule
        Configuration & MuSR & GSM8K \\
        \midrule
        Single-train, single-test (Qwen-only) & 53.6\% & 93.1\% \\
        Single-train, multi-test (cross-pool) & \textbf{69.5\%} & 90.5\% \\
        Multi-train, multi-test (in-domain) & 67.7\% & \textbf{97.0\%} \\
        \bottomrule
    \end{tabular}
\end{table}

\subsection{Confounding Analysis}
\label{app:confounding}

We screen all five datasets for model-identity confounding by computing two diagnostics: (i)~\emph{per-model pick distribution} (does the scorer disproportionately select one generator?) and (ii)~\emph{cross-model correct energy spread} (do correct candidates from different generators receive similar energies?).
\Cref{tab:confounding_screen} summarises the results.

\begin{table}[h!]
    \caption{Confounding screening across all datasets. Pick distribution shows the fraction of test problems where each generator's candidate is selected. Energy spread is the standard deviation of mean correct energy across generators (lower = less confounded).}
    \label{tab:confounding_screen}
    \centering
    \resizebox{0.9\textwidth}{!}{%
    \begin{tabular}{@{}lccccc@{}}
        \toprule
        Dataset & Qwen-2.5-7B & LLaMA-3.1-8B & Gemma-4-26B & GPT-OSS-20B & Energy spread \\
        \midrule
        GSM8K & 26\% & 24\% & 25\% & 25\% & 0.04 \\
        MuSR & 36\% & 23\% & 18\% & 23\% & 0.18 \\
        TravelPlanner & 22\% & 28\% & 26\% & 24\% & 0.24 \\
        K\&K & 28\% & 28\% & 27\% & 18\% & 0.11 \\
        \midrule
        TACO (pre-DFR) & 1\% & 1\% & \textbf{94\%} & 4\% & \textbf{6.67} \\
        TACO (post-DFR) & 22\% & 24\% & 34\% & 20\% & 0.34 \\
        \bottomrule
    \end{tabular}%
    }
\end{table}

Four datasets show near-uniform pick distributions and energy spreads below 0.25, indicating no model-identity confounding.
On MuSR, the scorer picks the \emph{weaker} model (Qwen, 51\% accuracy) most often (36\%), the opposite of what fingerprinting would produce.
On K\&K, despite the most extreme accuracy split in our study (Gemma/GPT-OSS at 99\% vs.\ Qwen/LLaMA at 15\%), picks remain uniform and the energy spread is just 0.11.

TACO is the sole outlier: the initial scorer selected Gemma on 94\% of test problems with an energy spread of 6.67, indicating severe model-identity confounding.
The six-phase analysis below details how this was detected, diagnosed, and mitigated.

\paragraph{TACO: detailed analysis.}
A six-phase confounding analysis on the TACO multi-model scorer revealed that the initial model exploited a model-identity shortcut rather than genuine reasoning-quality discrimination.                 
                  
\paragraph{Phase 1-2: Detection.} 
The scorer selected a Gemma-4-26B candidate on 83/88 test problems (94.3\%). A naive ``always pick random Gemma'' baseline achieved 89.8\% pass@1, leaving only +1.1\,pp of genuine quality signal. Correct solutions from different models received dramatically different energies: Gemma correct at $-$4.47 vs.\ LLaMA correct at $+$2.20, a 6.67-point gap for identical correctness labels.

\paragraph{Phase 3: Style normalisation ablation.}
Progressive removal of formatting features (headers, LaTeX, IO normalisation, comments) collapsed both the cross-model energy gap (6.67 $\to$ ${\sim}$1.0) and the within Gemma correct/incorrect gap (2.41 $\to$ 0.55), confirming that quality and identity features are carried by the same tokens.

\paragraph{Phase 4: Causal style-swap test.} 
For 40 problems with correct Gemma solutions, three other models rewrote each solution in their own style with variable names, algorithm, and logic locked. Execution-verified identical code received +1.95 higher energy on average when presented in another model's formatting (LLaMA: +2.66, GPT-OSS: +2.20, Qwen: +0.97). This is causal proof of style bias~\citep{geirhos_2020_shortcut}.

\paragraph{Phase 5-6: Decomposition and case studies.} 
Code-only scoring retained most of the model-identity gap (3.6-point spread), while explanation-only scoring collapsed to 62.5\% pass@1. The identity signal lives primarily in code patterns (IO methods, function structure), not explanation text. All 5 non-Gemma picks by the original scorer occurred on EASY problems; on 7 problems the scorer picked an incorrect Gemma solution despite correct alternatives being available.

\paragraph{Mitigation: DFR.}Last-layer retraining (DFR;~\citealp{kirichenko_2023_dfr}) on group-balanced data (4 models $\times$ 2 labels, capped at 396 per cell) retrains only the $K{=}5$ energy heads (0.4\% of parameters) while freezing the backbone and LoRA adapters. This collapses the cross-model correct energy spread from 6.67 to 0.34 (95\% reduction) while retaining 88.6\% pass@1 with picks distributed uniformly across generators (22/24/34/20\%). The deconfounded scorer improves pass@1 on HARD (+13.3\,pp) and UNKNOWN (+14.4\,pp) problems and is robust to two-pass re-pooling (88.6\% $\to$ 88.6\%, vs.\ pre-DFR catastrophic collapse 92.0\% $\to$ 2.3\%).

\section{Broader Impact}
\label{app:impact}
\vspace{-5pt}

This paper presents work whose goal is to advance machine learning, specifically the verification of structured outputs from large language models. We outline expected positive and negative societal impacts.

\paragraph{Positive impacts.} Uncertainty-aware verification of structured LLM outputs has beneficial applications in safety-critical domains where undetected errors propagate silently: \emph{automated planning} (travel itineraries, logistics, scheduling) where a single budget or routing violation invalidates the entire plan; \emph{code generation} where edge-case failures can cause silent data corruption or service outages; \emph{multi-step mathematical and scientific reasoning} where intermediate errors compound; and \emph{decision support} in regulated settings (legal, medical, financial) where outputs must satisfy domain rules verifiable in closed form. The decomposition $E = \muq + \lambda \cdot \Econstraint$ is auditable by design: $\Econstraint$ is a deterministic checker of explicit task constraints, and the distributional $(\muq, \sigmaq)$ exposes \emph{when the verifier itself is uncertain}, enabling principled abstention rather than confidently wrong outputs. Compared to scaling generators, a small specialised verifier on a fixed candidate pool is also more energy-efficient and democratises access to reliable structured reasoning without 100B+-parameter inference.

\paragraph{Negative impacts and risks.} Overreliance on automated verification carries three risks. \emph{(i) Spurious shortcuts.} Our confounding analysis (\cref{sec:exp:q4},~\cref{app:confounding}) demonstrates that scorers can learn high-confidence but content-irrelevant signals (the TACO formatting shortcut), producing reliable-looking but unreliable outputs. Without the kind of cross-dataset auditing we report, such shortcuts go undetected. \emph{(ii) Distribution shift.} A scorer trained on one candidate distribution may degrade silently when deployed against differently-styled generations or out-of-domain inputs (illustrated by the GSM8K$\to$MATH-500 below-random transfer; see \cref{tab:full_transfer}). \emph{(iii) Automation bias.} Plans, code, or proofs that pass an automated checker may be accepted with less human review than warranted, particularly when $\sigmaq$ is uncalibrated (e.g., MuSR $\sigma$-AUROC = 0.44; see Limitations in \cref{sec:exp:results}).

\paragraph{Mitigations.} We recommend: (a) running confounding diagnostics (cross-model pick distributions and energy spreads) at deployment time, with last-layer retraining (DFR;~\citealp{kirichenko_2023_dfr}) when shortcuts surface; (b) treating $\sigmaq$ as a regime-aware signal rather than a universal abstention threshold (\cref{sec:exp:q3} shows it is task-dependent); (c) preserving human review for high-stakes deployments and using $\sigma$-triage to route uncertain cases to humans rather than auto-accepting them; (d) documenting candidate-pool composition and constraint definitions when releasing checkpoints, since both interact materially with verifier behaviour. The released artefact in this work is a verifier/reranker (not a generative model), which substantially limits direct misuse risk relative to released LLMs.

\paragraph{Scaling $\Econstraint$.} The decomposition $E = \muq + \lambda \cdot \Econstraint$ requires a domain expert to author each constraint term, which is the principal scalability bottleneck of the framework. Three directions seem viable. \emph{(a) LLM-generated specifications:} a frontier model can be prompted to extract executable constraints from a natural-language task specification, with the EBM then ranking candidates against the auto-generated checker; the verifier inherits the spec-extractor's errors, but an exact deterministic constraint can be cheaper to author than to learn. \emph{(b) Learned constraint heads:} a small model trained on labelled violation data can approximate $\Econstraint$ for domains where authoring is hard but labelled examples exist (CI logs for code, structured grading rubrics for student work, regulatory checklists in compliance settings). \emph{(c) Constraint DSLs:} domain-specific languages for stating constraints, analogous to integrity constraints in relational databases, lower the authoring burden without removing the human in the loop. We treat these as engineering frontiers rather than methodological gaps; the propositions of \cref{sec:method:quality} hold regardless of how $\Econstraint$ is obtained, as long as it is deterministic at inference time.



\end{document}